\pdfoutput=1

\documentclass[11pt]{article}

\usepackage{acl}

\usepackage{booktabs}
\usepackage{url}
\usepackage{times}
\usepackage{latexsym}
\usepackage{tcolorbox}
\usepackage{subcaption}
\usepackage{amsfonts}
\usepackage[T1]{fontenc}
\usepackage{pgf-pie}

\usepackage[utf8]{inputenc}

\usepackage{microtype}

\usepackage{inconsolata}
\usepackage{amsmath}
\usepackage{graphicx}
\usepackage{multirow}
\usepackage{booktabs}
\usepackage{bbding}
\usepackage{MnSymbol}
\usepackage{xcolor}
\usepackage{colortbl}
\usepackage{enumitem}
\usepackage{placeins}
\usepackage{svg}
\usepackage{pifont}
\usepackage{lipsum}  
\usepackage[normalem]{ulem}
\usepackage{enumitem} 
\usetikzlibrary{patterns}
\usepackage{algorithm}
\usepackage{algpseudocode}

\usepackage{siunitx}
\usepackage{titletoc}

\newcommand\corpusname{\textbf{\texttt{EtiCor}}}
\newcommand\newcorpusname{\textbf{\texttt{EtiCor++}}}

\title{EtiCor++: Towards Understanding Etiquettical Bias in LLMs}

\author{
Ashutosh Dwivedi\thanks{Equal Contribution}
\qquad
Siddhant Shivdutt Singh\footnotemark[1]
\qquad
Ashutosh Modi
\\ 
        Indian Institute of Technology Kanpur (IIT Kanpur)
\\
  \texttt{\{ashutoshd20,siddhss20\}@iitk.ac.in} \qquad \texttt{ashutoshm@cse.iitk.ac.in}  
}

\begin{document}
\maketitle


\setlength{\abovedisplayskip}{0pt}
\setlength{\belowdisplayskip}{0pt}
\setlength{\abovedisplayshortskip}{0pt}
\setlength{\belowdisplayshortskip}{0pt}

\begin{abstract}
In recent years, researchers have started analyzing the cultural sensitivity of LLMs. In this respect, Etiquettes have been an active area of research. Etiquettes are region-specific and are an essential part of the culture of a region; hence, it is imperative to make LLMs sensitive to etiquettes. However, there needs to be more resources in evaluating LLMs for their understanding and bias with regard to etiquettes. In this resource paper, we introduce \textbf{EtiCor++}, a corpus of etiquettes worldwide. We introduce different tasks for evaluating LLMs for knowledge about etiquettes across various regions. Further, we introduce various metrics for measuring bias in LLMs. Extensive experimentation with LLMs shows inherent bias towards certain regions.  
\end{abstract}

\section{Introduction} \label{sec:intro}

\noindent In recent times, Large Language Models (LLMs) have shown drastic improvements across almost all NLP tasks involving language understanding and generation \cite{chang2024survey,patra-etal-2023-beyond,10.1145/3554727,zhong2024evaluationopenaio1opportunities}, resulting in wide-spread adoption in real life applications such as using LLM as personal digital assistants where the LLM is used for querying about various kinds of information including those related to cultural aspects of human societies. Consequently, the NLP research community has recently started focusing on evaluating and improving cultural understanding (and possible biases) of LLMs \cite{hershcovich-etal-2022-challenges,abrams-scheutz-2022-social,li2024culturepark}. It has resulted in the need to develop new culture-centric tasks and datasets. Culture is a multi-faceted topic and has been studied in the NLP community via various proxies  \cite{adilazuarda-etal-2024-towards}. One aspect of culture is Etiquettes.\footnote{In this work, we follow the previous definition of etiquette as defined in \citet{dwivedi-etal-2023-eticor}: a set of social norms/conventions or rules that tell how to behave in a particular social situation.} Etiquettes can be generic (common across the majority of societies/regions) as well as localized (specific to a society/region). LLMs have been trained on almost the entire internet's data \cite{villalobosposition} and have very likely picked up information about etiquettes in various societies. However, it remains to be evaluated if LLMs are able to understand intricate and subtle differences in social norms across cultures and are possibly biased towards certain cultures.  Since LLMs are increasingly being used for seeking information, potential etiquettical biases can have detrimental consequences for the user. Hence, there is a need for evaluation and understanding of inherent biases. In this resource paper, we attempt to achieve this goal. In a nutshell, we make the following contributions:

\begin{figure}[t]
    \centering
    \includegraphics[scale=0.32]{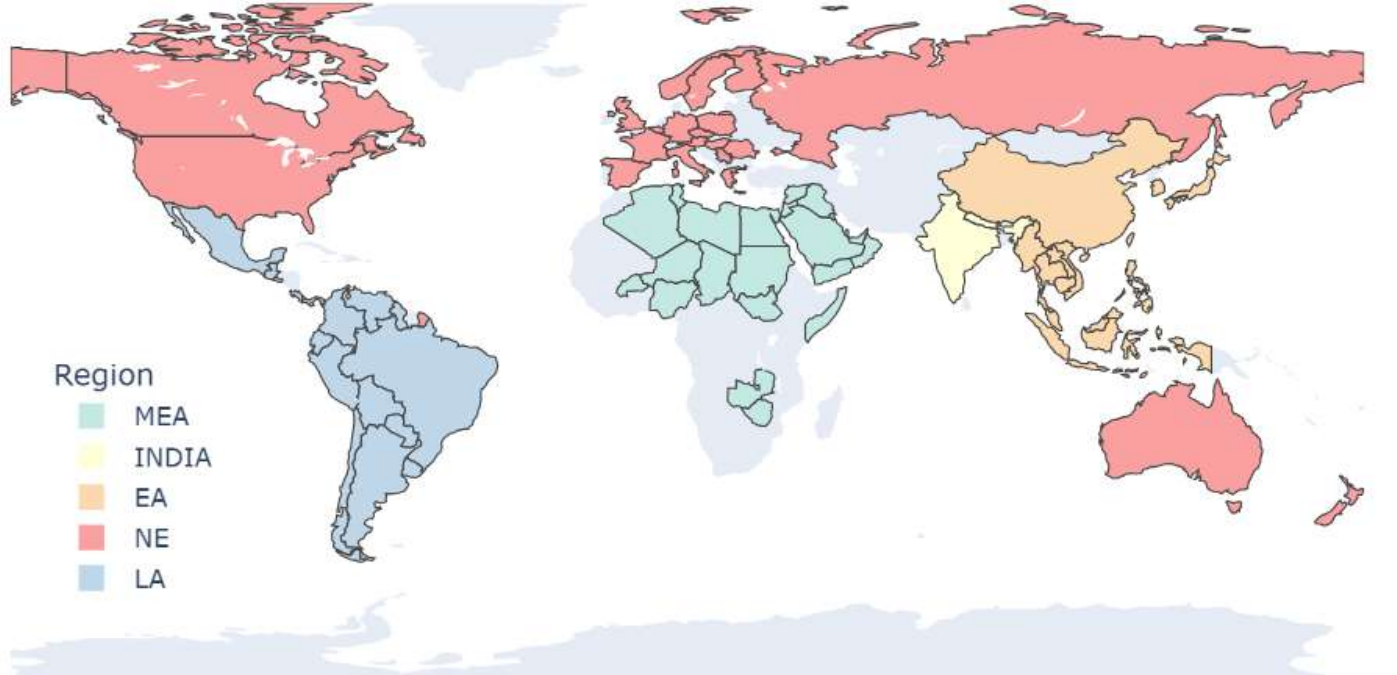}
    \vspace{-1mm}
    \caption{Regions covered under \newcorpusname}
    \label{fig:eticor-plus-distribution}
    \vspace{-4mm}
\end{figure}

\begin{itemize}[nosep,noitemsep,leftmargin=*]
\item In this paper, we introduce a new language resource:  \newcorpusname, a large English corpus of $48K$ etiquettes that cover the majority of regions across the globe as shown in Fig. \ref{fig:eticor-plus-distribution}. 
\item We perform an in-depth analysis of \newcorpusname. Though etiquettes vary from region to region, there are some commonalities. We develop an algorithm (Algorithm \ref{algo:corr}) for measuring the correlation between etiquettes of different regions to check for similarities and differences. 
\item In addition to the existing task of  Etiquette Sensitivity \cite{dwivedi-etal-2023-eticor}, we propose two new tasks (Region Identification and Etiquette Generation) to evaluate LLMs for the understanding of etiquettes across regions. 
\item We propose seven new algorithmic metrics (\S\ref{sec:method}) for measuring Etiquettical bias in LLMs: Preference Score, Bias For Region Score, Pairwise Regions Bias Score, Generation Alignment Score, Odds Ratio, and two variants of Incremental Option Testing. 
\item We conduct an extensive set of experiments on five popular LLM models (Llama3.1,  Phi-3.5-mini, Gemma2, Gemini, and GPT-4o); our experiments show that LLMs tend to prefer certain regions more than others when it comes to social norms. We release dataset and code via GitHub: \url{https://github.com/Exploration-Lab/Eticor-Plus-Plus}
\end{itemize}

\section{Related Work} \label{sec:related}

\noindent \textbf{Culture-centric Research in NLP Community:} With the aim to deploy NLP technologies (e.g., LLMs) in human societies, recent research in the NLP community has focused on ethics and culture-centric techniques and models \cite{adilazuarda-etal-2024-towards,ziems-etal-2023-normbank,agarwal-etal-2024-ethical}. Various works have been proposed covering social reasoning  \cite{DBLP:journals/corr/abs-2110-07574}, cross cultural understanding \cite{pandey-etal-2025-culturally}, social dimensions \cite{hershcovich-etal-2022-challenges}, CultureLLM  \cite{DBLP:journals/corr/abs-2307-07870}, cultural corpora \cite{10.1145/3543507.3583535,cao-etal-2024-bridging,ammanabrolu-etal-2022-aligning}, LLM alignment \cite{alkhamissi-etal-2024-investigating}, and inter alia. Due to space constraints we provide more details in App. \ref{app:related-work}. 

\noindent \textbf{Comparison with \corpusname:} We are inspired by \citet{dwivedi-etal-2023-eticor},  where the authors create a corpus of etiquettes (\corpusname) from major regions of the world and propose the task of Etiquette sensitivity. Since \corpusname\ is available under open-source license, our work \newcorpusname\ takes \corpusname\ as the starting point (removes some of the noisy samples by manually analyzing it) and extends \corpusname\ significantly from $35K$ etiquette text (each text roughly equivalent to a sentence) to $48K$. We have included many more diverse cultures around the world, such as the Aborigines of Australia, Maori in New Zealand, China, Russia, and southern parts of Africa, which were missing in the previous dataset. The corpus coverage per region has been expanded from a set of few countries to several nearby countries. 
The consensus of joining the countries to this list was based on the idea of common inclusion. We perform an in-depth analysis and propose a new algorithm for measuring the correlation between etiquettes belonging to different regions. Previous work had only one task for measuring etiquettical knowledge; we have included new tasks and metrics along with evaluation using the latest LLMs. 

\noindent \textbf{Measuring Bias and Stereotypes:} There has been extensive research on measuring biases and stereotypes in deep models and LLMs \cite{gallegos-etal-2024-bias,shrawgi-etal-2024-uncovering}. This paper highlights only the relevant works (details in App. \ref{app:related-work}). Researchers have addressed  stereotypical biases in models \cite{10.1037/pspa0000046,cao-etal-2022-theory,nadeem-etal-2021-stereoset,nangia-etal-2020-crows,jha-etal-2023-seegull,10.5555/3666122.3666316,das-etal-2023-toward,palta-rudinger-2023-fork}, persona bias \cite{wan-etal-2023-personalized}, effect of cultural bias on NLU \cite{wan-etal-2023-kelly,huang-yang-2023-culturally}.

\begin{tcolorbox}[width=\columnwidth,colback=gray!10,colframe=gray,sharp corners,title=Example,coltitle=white!150!black,size=small]
    \textbf{Sentence 1:} \texttt{Living with your parents after the age of 18 is considered a bad practice.} \\
    \textbf{Sentence 2:} \texttt{Young people choose to live with their parents even after the age of 18 and it's considered okay.} \\
    \textbf{Sentence 3:} \texttt{Independent living after a certain age is considered more appropriate in this culture.} 
\end{tcolorbox}
 
\noindent \textbf{Motivation for New Metrics:} Existing works have very little coverage (mostly restricted to sentence-level semantic similarity) for evaluating the generative capabilities of LLMs in the context of culture and, in particular, in the context of etiquettes. Consider the example shown above. As per sentence similarity models (sentence-transformers/all-mpnet-base-v2), the first and second sentences are more similar (0.792) than the first and third (0.501), even though the first two convey opposite values. To take care of nuanced responses and their alignment, we propose new metrics inspired by the NLI task \cite{storks2019recent}. Most works focus on a sensitivity-based bias analysis where a model is evaluated for culture based on food, names, gender, or other proxies \cite{adilazuarda-etal-2024-towards}. We wanted to have a metric that quantifies the bias of an LLM in mapping etiquettes to cultures/regions to cover broader use cases mentioned in \S\ref{sec:method}.

\section{\newcorpusname} \label{sec:corpus}

\begin{table}[t]
\centering
\scriptsize 
\renewcommand{\arraystretch}{1.35} 
\resizebox{\columnwidth}{!}{ 
\begin{tabular}{>{\raggedright\arraybackslash}m{4.25cm} >{\centering\arraybackslash}m{1.2cm} >{\centering\arraybackslash}m{0.75cm} >{\centering\arraybackslash}m{1.0cm}}
\toprule
\textbf{Etiquette} & \textbf{Group} & \textbf{Region} & \textbf{Label} \\ 
\midrule
Observing seniority and rank are extremely important in business. & Business & EA & Positive \\ 
It’s not alright to eat beef in front of the people. & Dining & India & Positive \\ 
Hand your tip to the waiter, do not leave it on the table. & Dining & LA & Positive \\ 
A small burp signifies satisfaction. & Visits & MEA & Positive \\ 
Familiarize yourself with any posted visitor guidelines or rules before entering a European location. & Travel & NE & Positive \\ 
Women and men generally eat together. & Dining & MEA & Negative\\
The business meal is generally the time to make business decisions. & Business & LA & Negative\\
It's okay to stand with your hands on your hips while talking with someone. & Visits & EA & Negative\\
\bottomrule
\end{tabular}
}
\vspace{-3mm}
\caption{\newcorpusname\ corpus examples.} 
\label{tab:etiquette-examples}
\vspace{-4mm}
\end{table}

\newcorpusname\ contains $47,720$ region-specific etiquette texts in English. As done in previous work \cite{dwivedi-etal-2023-eticor} we intentionally do not have a multi-lingual corpus due to reasons related to maintaining compatibility across regions and the possibility of introducing biases during translation (see Limitations for details). We have categorized etiquettes into five regions. Table \ref{tab:etiquette-examples} shows examples of sentences belonging to different regions. Each region is sub-categorized into one of the 4 four social activities (Dining, Travel, Visits, Business). Further, each etiquette is assigned a label: ``Positive" (acceptable in the region) or ``Negative" (not acceptable in the region). Table \ref{tab:etiquette-type-distribution} shows the corpus statistics. We created \newcorpusname\ by scraping, manually cleaning, and refining content from authentic government websites and travel blogs/websites (details in App. \ref{app:corpus-creation}).

\noindent \textbf{Regions in \newcorpusname:} We categorize the etiquettes collected across the globe into five regions (East Asia (EA), Middle East and Africa (MEA), India Subcontinent (IN), Latin America (LA), and North America and Europe (NE) (also see Fig.~\ref{fig:eticor-plus-distribution}). Compared to \corpusname, the region names have also been updated since several new countries were added. Consequently, the region-wise categorization is different from \corpusname. Countries that share culture and several other aspects, such as religion, dining, and history, are brought under one region. For example, Russia is included in the North-America-Europe (NE) region due to similarities in dining habits and shared history. We also include some countries in a common region, even though they are geographically far away, such as European countries, Australia, and New Zealand. 
Note that social norms are very often common in geographically close countries. However, this was not the reason to club countries into one region (details in App.~\ref{app:corpus-creation}). Following regions are created:\\
\textbf{a) East Asia (EA):} This region includes Japan, Korea, Taiwan, China, and all the other Southeast Asian countries, e.g., Indonesia, Malaysia, Philippines, Thailand, Vietnam, etc. There is a significant overlap in these countries' cultural and social values; hence, to maintain harmony, they are in one region. Nevertheless, country information is maintained along with the etiquette. \\
\textbf{b) Middle East and Africa (MEA):} We studied the information collected for countries in the Middle East and Africa and excluded texts very niche to certain religious and tribal practices. It was done to maintain consistency across etiquettes in the MEA region. Africa could not be separated from the Middle East due to the lack of data, and a detailed study of the contrast of regions is required. Furthermore, the North Africa and Middle-East regions shared more cultures and practices than Southern Africa. Thus, we only included some Southern African countries with common etiquettes. In the future, once we have more data available, we plan to create a separate region (and sub-regions) for Africa. \\ 
\textbf{c) Indian Subcontinent (IN):} We created a separate region for India (and its neighboring countries) due to its vibrant sociocultural diversity. We also include Nepal in this region due to a high overlap in the social practices between the two countries. We use the terms India and Indian Subcontinent interchangeably.

\begin{table}[t]
\tiny
\centering
\renewcommand{\arraystretch}{1}
\setlength\tabcolsep{3pt}
\begin{tabular}{lccccc}
\toprule
\textbf{Region} & \textbf{\# Travel} & \textbf{\# Business} & \textbf{\# Visits}& \textbf{\# Dining} & \textbf{Total} \\
\toprule
\textbf{EA} & 1190 & 2960 & 4878 & 1100 & 10128\\
\textbf{MEA} & 1776 & 3448 & 6984 & 1300 & 13508\\
\textbf{IN} & 364 & 1104 & 2252 & 450 & 4170\\
\textbf{LA} & 1044 & 2058 & 3166 & 996 & 7264\\
\textbf{NE} & 1330 & 3548 & 6284 & 1488 & 12650\\ 
\midrule
\textbf{Total} &5704& 13118 & 23564 & 5334 & \textbf{47720}\\
\bottomrule
\end{tabular}
\vspace{-3mm}
\caption{Distribution of different etiquette types}
\label{tab:etiquette-type-distribution}
\vspace{-4mm}
\end{table}

\begin{algorithm}[h]
\small
\renewcommand{\algorithmicrequire}{\textbf{Input:}}
\renewcommand{\algorithmicensure}{\textbf{Output:}}
\caption{Inter-Region Correlation}\label{algo:inter-region-correlation}
\begin{algorithmic}
\Require \{$E_{i}^{(R_{j})} \forall i \in \{1, \ldots, n_{R_{j}} \}, j \in \{1, \ldots, 5 \}$ \} : Etiquettes from each of the five regions. 
\State $G \in \text\{Dining, Travel, Business, Visits\}$
\Ensure Corr($R_{j}, R_{k}$) $\forall j, k \in \{1, \ldots, 5 \}; j \ne k$
\State Start:
\State Initialize Corr($R_{j}, R_{k}$) = [5][5]
\State Calculate embedding for each etiquette using SBERT:  
\State \qquad $S_{E_{i}^{(R_{j})}} = SBERT (E_{i}^{(R_{j})}) $
\For{$j$ in \{1, \ldots, 5 \}}
\For{$i$ in $\{1, \ldots, n_{R_{j}} \}$}
\State Initialize $CorrList(n_{R_{j}}, 5) = [ ] [ ]$
\For{$k$ in \{1, \ldots, 5 \}; $k \ne j$}
\State Initialize  $SimList^{(j)}(i) = [ ]$
\For{$l$ in $\{1, \ldots, n_{R_{k}} \}$}
\State \textbf{if} $G(E_{i}^{(R_{j})}) \ne G(E_{l}^{(R_{k}})$ 
\State \qquad \textbf{continue}
\State $sim(i,l) =  CosSim(S_{E_{i}^{(R_{j})}}, S_{E_{l}^{(R_{k})}}) $
\State append $sim(i,l)$ in $SimList^{(j)}(i)[]$
\EndFor
\State $m = argmax (SimList^{(j)}(i))$
\State Determine relationship $(\mathcal{R})$ using MNLI model
\State $\mathcal{R} \in $  \{Supportive, Contrastive\} $\in \{ +1, -1 \}$
\State $\mathcal{R}(i, m) = \ $RoBERTa-MNLI$(E_{i}^{(R_{j})}, E_{m}^{(R_{k})})$
\State $Corr(i^{(j)}, m^{(k)}) = \mathcal{R}(i, m) * sim(i, m)$
\State Append $Corr(i^{(j)}, m^{(k)})$ in $CorrList[i][]$
\EndFor
\EndFor
\State Corr($R_{j}, R_{k}$) = $mean(CorrList[:][k])$
\State Append Corr($R_{j}, R_{k}$) in Corr[j, k]
\EndFor
\end{algorithmic}
\vspace{-1mm}
\label{algo:corr}
\end{algorithm}

\noindent \textbf{d) Latin America (LA):} This region has a large geographical area covering diversity in etiquettes. After an in-depth study of the cultural similarities, Cuba and Colombia are included in this region. \\
\textbf{e) North America and Europe (NE):} This region, due to prominent social and cultural commonalities, includes the U.S.A., Canada, Australia, New Zealand, and Russia. Even though these countries are geographically apart, they have high cultural similarity as well as historical alignment, hence these are clubbed together.  

\begin{figure}[t]
    \centering
    \includegraphics[scale=0.25]{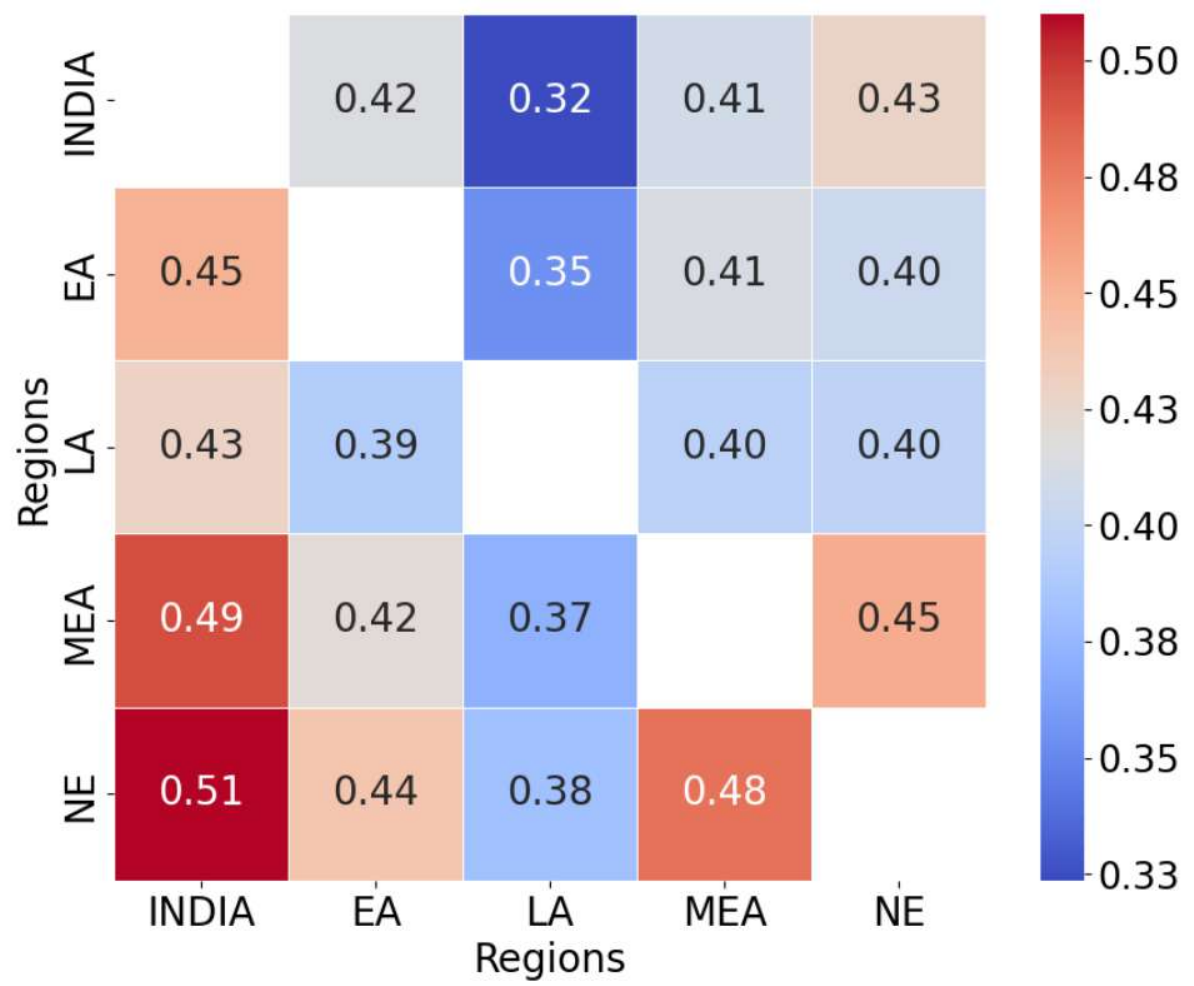}
    \vspace{-2mm}
    \caption{Region wise Correlation}
    \label{fig:inter-region-corr}
    \vspace{-6mm}
\end{figure}

\noindent\textbf{Inter-Region Analysis:} We analyzed the correlation between etiquettes across five regions to study the similarities and differences in social norms globally. Given the complex sociocultural nature of etiquettes, it is not straightforward to measure correlation; however, in this paper, we adopted the simplest possible approximate measure based on semantic similarity and NLI (also see Limitations section). We propose Algorithm \ref{algo:corr}. Note, as explained above, this algorithm only serves as a proxy for measuring correlation between regional etiquettes. In Algorithm \ref{algo:corr}, for finding the correlation between a region with others, first the similarity (using SBERT model \cite{reimers-gurevych-2019-sentence}) between an etiquette is compared with etiquettes belonging to the same group (dining, travel, business, and visits); 
next among all the etiquettes of other regions (with which similarities were calculated), the one with maximum value is selected and compared (via RoBERT-MNLI model \cite{liu2019roberta}) with the original etiquette to find out if it supports it or contradicts it. We wanted to use a pre-trained off-the-shelf NLI model for our experiments. Consequently, we went with the most readily available model: RoBERTa-MNLI. Correlation is approximated by taking the product of similarity and NLI score. The process is repeated for each of the etiquettes in the original region. Fig. \ref{fig:inter-region-corr} shows the correlation between $R_{j}$ and $R_{k}$ with $R_{j}$ at x-axis. Note (as can be inferred from Algorithm \ref{algo:corr}), Corr($R_{j}, R_{k}$) $\ne$ Corr($R_{k}, R_{j})$. The important point to note here is that the correlations are not diagonally symmetric because the number of data points in each region is not same. Hence, a region R1 can have a high overlap with the R2 region, but the percentage of points considered could make up only $20\%$ of the total points of R2, thus enabling R2 to have a higher match with other regions. It also gives an idea of the global distribution of data points available on the internet and its effect on LLM during training. As can be observed, the LA region has the least similarity with the rest of the regions (possibly because of geographical distances) and the highest similarity with Europe (since Europeans colonized it). IN has large similarities with other regions (possibly because they were colonized at various times in history). We also calculate the group-wise correlation between regions (see App. \ref{app:grp-corr}). We also provide statistics related to General Etiquettes in App. \ref{app:general-etiquette}. 

\section{Tasks and Bias Metrics} \label{sec:method}

We use \newcorpusname\ to check LLMs for cultural bias. For this, we created various tasks and metrics for measuring bias, as outlined below.

\noindent\textbf{Etiquette Sensitivity (ES) Task:} This task is similar to the one introduced in \citet{dwivedi-etal-2023-eticor}. Given an etiquette, the task is to predict whether the etiquette is acceptable or unacceptable for a region. We evaluate LLMs via zero-shot setting (App. \ref{app:prompts} provides the prompt). This task is useful as we need the models to be sensitive to different cultures and not discriminate against any of them. The models should not deem some cultural values acceptable while others are unacceptable. ES is measured using the standard metric of Accuracy and F1 score. 

\noindent\textbf{Region Identification (RI) Task:} This newly introduced task aims to test if a model can correctly identify the region corresponding to an etiquette. The model is provided with an etiquette text and asked to identify the region from a list of regions (see App. \ref{app:prompts} for the prompt). We created this task by keeping the following use cases in mind. Let's say that a person asked the LLM to suggest a gift for their friend's wedding. However, the LLM is unaware of the friend's cultural belonging. Suppose the model responded with ``\texttt{You can gift them white flowers, as it represents purity and peace}'' but then you respond with ``\texttt{In our culture, White is an ominous color for us. Please suggest a different gift.}'' Now, the model can actually guess what culture is involved here (e.g., East Asian) and respond accordingly. 
There are many other use cases, such as asking the model to assist you in writing a speech at somebody's funeral or promotion, which can involve etiquette regarding first and last names, etc. We devise three metrics to evaluate a model. 

\noindent\textbf{\textit{1. \underline{Preference Score ($\mathbf{PS(R)}$)}}} for a region $\mathbf{R} \in \text{\{EA, IN, MEA, LA, NE\}}$ calculates how often the model prefers to select the region across all etiquettes in the corpus, i.e., 
\begin{align*}
\mathbf{PS(R)} = \frac{\sum_{i=1}^{N} \mathbb{I}_{\mathbf{R} == RI(E_{i})}}{N}    
\end{align*}

\noindent where, $\{\mathbb{I}_{a} = 1\ \text{if} \ a = \text{\texttt{True}}\}$ is the indictor function, $N$ is total number of etiquettes in the corpus and $RI(E_{i})$ is the answer generated by the model for the RI task query. A higher value of $\mathbf{PS(R)}$ than expected is indicative of a model's bias towards a region. The expected value for each region is their share in the actual data distribution (NE - 26.50\%, IN - 8.73\%, EA - 21.22\%, LA - 15.22\%, MEA - 28.30\%). To estimate this deviation, we calculate 
\begin{align*}
(\mathbf{PS(R)\%} - \mathbf{D(R)}), 
\end{align*}
\noindent where $\mathbf{D(R)}$ is the percentage share of the region $\mathbf{R}'s$ data in the whole dataset. We also calculate standard deviation ($\sigma_{PS(R)}$) for each model (\S\ref{app:sec-psr}). 

\noindent\textbf{\textit{2. \underline{Bias For Region Score ($\mathbf{BFS(R)}$)}}} for a region $\mathbf{R}$ is calculated by iterating over all etiquettes not in $\mathbf{R}$ and checking how often the RI query returns $\mathbf{R}$ as the answer, i.e., 
\begin{align*}
\mathbf{BFS(R)} = \frac{\sum\limits_{\mathbf{R'} \ne \mathbf{R}} \sum\limits_{i = 1}^{N_{\mathbf{R'}}} \mathbb{I}_{RI(E_{i}) == \mathbf{R}} }{\sum_{\mathbf{R'} \ne \mathbf{R}} N_{\mathbf{R'}} }
\end{align*}
Using this score, we look at the defaulting behavior of the LLMs. We are trying to quantify that whenever a model is wrong about the region of an etiquette, what region does it prefer in those cases. A high $\mathbf{BFS(R)}$ score indicates model bias for a region. Similar to $\mathbf{PS(R)}$, we calculate standard deviation ($\sigma_{BFS(R)}$) for each model (\S\ref{app:sec-bfsr}). 

\noindent\textbf{\textit{3. \underline{Pairwise Regions Bias Score ($\mathbf{BSP(R, R')}$)}}} Given RI predictions for etiquettes in region $\mathbf{R}$, $\mathbf{BSP(R, R')}$ assess how often the incorrect predictions are confused for region $\mathbf{R'}$, i.e.,
\begin{align*}
\mathbf{BSP(R, R')} = \frac{\sum\limits_{i = 1}^{N_{\mathbf{R}}} \mathbb{I}_{RI(E_{i}) == \mathbf{R'}}  }{ \sum\limits_{i = 1}^{N_{\mathbf{R}}} \mathbb{I}_{RI(E_{i}) \ne \mathbf{R}} }
\end{align*}
BSS is not a symmetric metric (higher the score more the bias).

\noindent\textbf{Etiquette Generation (EG) Task:}
Restricting the LLM to specified options, as in previous tasks, might prevent us from observing the generational biases of the model. We propose the Etiquette Generation (EG) task where a model is provided with an etiquette of one region in one context (e.g., group or etiquette type) and is asked to generate an etiquette for the other regions in the same context (see App. \ref{app:prompts} for the prompt template). We propose this task as we want the real-world LLM response to be non-stereotypical and non-contradictory. A possible use-case may involve a person (belonging to a region) simply wanting to know about the traditions, norms, values, and etiquette of any other culture in a particular context. We want to qualitatively and quantitatively assess these properties. We propose two new metrics (Generation Alignment Score (GAS) and Odds Ratio):

\noindent \textbf{\textit{1. \underline{Generation Alignment Score (GAS):}}} For all the generated etiquettes for a particular region $\mathbf{R}$, we would like to measure the alignment and consistency of generated responses for other regions. For this, we first calculate the embeddings for each generated etiquette using \textit{sentence-transformers/all-mpnet-base-v2} model \cite{reimers-2019-sentence-bert,DBLP:conf/nips/Song0QLL20} and then filter out etiquettes that have similarity less than a threshold (selected as 0.55 via initial experiments). However, this also resulted in etiquette that had contradictory stances being selected. Consequently, we used Natural Language Inference (NLI) to calculate the entailment and contradiction scores (a threshold of 0.90 was used to filter out). The GAS score is defined as: 
\begin{align*}
\frac{\#entailment}{\#entailment + \#contradictions}
\end{align*}
GAS helps us gauge the robustness and confidence of the model. GAS score lies between $0$ (worst) and $1$ (best).  

\noindent \textbf{\textit{2. \underline{Odds Ratio:}}} Inspired by the work of \citet{DBLP:conf/acl/NaousRR024}, we apply the Odds Ratio test to identify the dominating themes of the generated etiquettes. In particular, we analyze frequent Nouns, Verbs, and Adjectives in the responses generated for each pair of regions. This qualitative metric aims to investigate the generation of stereotypes for certain regions.  

\noindent\textbf{Incremental Option Testing:} The Region Identification task involves providing a query with one correct and a set of incorrect choices. However, it does not provide the means to evaluate the stability and confidence of the model about a set of choices. We propose the task of \textbf{Incremental Option Testing} for this purpose. We intend to create metrics that also map the stability of the model concerning the etiquette on which they are making the decisions. It helps us to understand the randomness they might exhibit when new data is presented and how it changes their decisions, ultimately resulting in changes in their bias in light of new information. In this task, a query is posed to the model along with options to select an answer (MCQA style). Initially, two options are provided and the model's response is recorded. Subsequently, the same query is posed again but with the addition of one more choice. Again, the model's response is recorded. The consistency within the sequence of predictions made by the model is observed. Algorithm \ref{algo:incremental} gives details. We consider two possible types of increments.

\begin{algorithm}[h]
\small
\renewcommand{\algorithmicrequire}{\textbf{Input:}}
\renewcommand{\algorithmicensure}{\textbf{Output:}}
\caption{Incremental Option Testing}
\label{algo:incremental-option-testing}
\begin{algorithmic}
\Require $\mathcal{Q} = \{Q^{(i)} \mid i = 1, \ldots, |\mathcal{Q}|\}$: Set of questions
\Ensure Choices, $\{C_{j}^{(i)} \mid i = 1, \ldots, |\mathcal{Q}|; j = 1, \ldots, m \}$: Model predictions for each question and iteration $(j)$. Given $(m+1)$ total number of options (regions).
\For{$i = 1, \ldots, |\mathcal{Q}|$} 
    \State Present question $Q^{(i)}$
    \State Present initial options $[ O_{1}^{(i)}, O_{2}^{(i)} ]$
    \State Let the model’s predicted choice be $C_{1}^{(i)}$
    \State Initialize $j = 2$ 
    \While{additional options remain to be tested}
        \State Introduce a new option $O_{j+1}^{(i)}$
        \State Query the model, yielding choice $C_{j}^{(i)}$
        \State Increment $j \gets j + 1$
    \EndWhile
    \State Record set of predictions $\{C_{j}^{(i)} \mid j = 1, \ldots, m\}$
\EndFor
\end{algorithmic}
\label{algo:incremental}
\end{algorithm}
\vspace{-2.5mm}


\noindent\textbf{\underline{\textit{1) Correct Option at the Start Increment:}}} In this method, the correct choice is introduced initially at index 0, and subsequently incorrect choices are introduced (in decreasing order of correlation based on Fig. \ref{fig:inter-region-corr}). The expected behavior of the model is to select the correct option at the first choice and not waiver from this decision even when new options are added to the list (prompt in App. \ref{app:prompts}). This variant is evaluated using two metrics:
\noindent \underline{\textit{Accuracy:}} We have the accuracy of the models for each set of options for a particular etiquette. The general trend shows a decline in accuracy with increase in number of options (\S\ref{sec:experiment_results}). Furthermore, some models perform decently in the initial step which suggest that given the limited  number of choices and possibility of their bias source lacking, they will have high accuracy (\S\ref{sec:experiment_results}).
\noindent \underline{\textit{Distancing:}} It measures distance between true and predicted choice (see App. \ref{app:sec-distancing} Algorithm \ref{algo:distancing} for calculation details). It gauges the increase in bias of the model as more choices are presented. An increase in magnitude of negative score states that the model is moving toward biased opinion and a sharper fall indicates a greater tendency to move towards the least possible options. We want the model to be as close to zero as possible. 

\noindent\textbf{\underline{\textit{2) Correct Option at the End Increment:}}} In this method, the choices are introduced one at a time, with the correct option introduced at the end. The incorrect options are added in the increasing order of region-wise correlation (Fig. \ref{fig:inter-region-corr}).  The typical behavior expected from the model is to pick the newest option from the choices. This approach is evaluated with three metrics: 

\noindent \underline{\textit{a) Closeness:}} This metric help us to understand how much is the bias of a model near the optimal value. Algorithm \ref{algo:closeness} gives the details of the calculation. 

\noindent \underline{\textit{b) Consistency:}} A consistency score determines the direction of the decision the model is making under the change of available information. Through this score we try to measure how consistent the models are when they are probed for the etiquette.
\begin{algorithm}[h]
\small
\renewcommand{\algorithmicrequire}{\textbf{Input:}}
\renewcommand{\algorithmicensure}{\textbf{Output:}}
\caption{Closeness Metric Calculation}\label{algo:closeness-calculation}
\begin{algorithmic}
\Require 
    \noindent $N$: Number of questions. \\
    \noindent $M$: Number of choice iterations (ITRs). \\
    \noindent $\{C_i^{(j)}\}$: Choice for question $Q_i$ at ITR $j$, $\forall i \in \{1, \ldots, N\}$, $\forall j \in \{0, \ldots, M-1\}$. \\
    \noindent $\{O_j\}$: Latest Options introduced at each ITR $j$, $\forall j \in \{0, \ldots, M-1\}$.
\Ensure $\{\text{Closeness}^{(j)}\}$: Closeness value for each phase $j$, $\forall j \in \{0, \ldots, M-1\}$.
\For{$j$ in $\{0, \ldots, M-1\}$}
    \For{$i$ in $\{1, \ldots, N\}$}
        \State Initialize score $S_i^{(j)} = 0$
        \If{$C_i^{(j)} = O_j$}
            \State $S_i^{(j)} = 0$
        \ElsIf{$C_i^{(j)} = O_{j-1}$}
            \State $S_i^{(j)} = -1$
        \ElsIf{$C_i^{(j)} = O_k \ \text{where} \ k < j-1$}
            \State $S_i^{(j)} = -2$
        \EndIf
    \EndFor
    \State Calculate $\text{Closeness}^{(j)} = \frac{1}{N} \sum_{i=1}^{N} S_i^{(j)}$
\EndFor
\State \textbf{Return} $\{\text{Closeness}^{(j)} \ \forall j \in \{0, \ldots, M-1\}\}$
\end{algorithmic}
\label{algo:closeness}
\end{algorithm}
Based on Algorithm \ref{algo:closeness}, 
\begin{align*}
\text{Consistency Score}^{(j)} = \frac{\sum\limits_{i=1}^N \mathbb{I}(S_i^{(j)} = -1)}{N},
\end{align*}
\noindent where, 
\begin{align*}
\mathbb{I}(S_i^{(j)} = -1) = 
\begin{cases} 
1 & \text{if } S_i^{(j)} = -1, \\
0 & \text{otherwise.}
\end{cases} 
\end{align*}
\begin{table*}[ht]
    \centering
    \tiny
    \begin{tabular}{c c c c c c c c c c c}
        \toprule
        \textbf{Region} & \multicolumn{2}{c}{\textbf{ChatGPT-4o}}  & \multicolumn{2}{c}{\textbf{Gemini-1.5}} & \multicolumn{2}{c}{\textbf{Llama-3.1}} & \multicolumn{2}{c}{\textbf{Gemma-2}} & \multicolumn{2}{c}{\textbf{Phi-3.5}} \\
        & \textbf{Acc.(\%)} & \textbf{F1-Score} & \textbf{Acc.} & \textbf{F1-Score} & \textbf{Acc.} & \textbf{F1-Score} & \textbf{Acc.} & \textbf{F1-Score} & \textbf{Acc.} & \textbf{F1-Score} \\
        \midrule
        NE    & 42.0 & 0.59 & 43.5 & \cellcolor{green!30}0.61 & 46.4$\pm$1.24 & \cellcolor{green!30}0.64$\pm$0.02 & 45.5$\pm$2.05 & \cellcolor{green!30}0.61$\pm$0.012 & 45.5$\pm$2.88 & 0.62$\pm$0.015 \\
        INDIA  & 44.0 & \cellcolor{green!30}0.61 & 41.0 & 0.58 & 45.2$\pm$2.03 & 0.62$\pm$0.025 & 43.2$\pm$1.94 & 0.60$\pm$0.009 & 45.9$\pm$2.86 & \cellcolor{green!30}0.63$\pm$0.024 \\
        EA & 41.1 & \cellcolor{red!30}0.58 & 37.6 & \cellcolor{red!30}0.54 & 41.3$\pm$1.22 & \cellcolor{red!30}0.60$\pm$0.021 & 40.7$\pm$1.76 & 0.57$\pm$0.005 & 42.2$\pm$1.62 & 0.59$\pm$0.015 \\
        LA   & 42.5 & 0.59 & 39.1 & 0.56 & 41.8$\pm$1.14 & 0.61$\pm$0.01 & 38.9$\pm$2.0 & \cellcolor{red!30}0.54$\pm$0.012 & 41.2$\pm$1.96 & \cellcolor{red!30}0.58$\pm$0.006 \\
        MEA & 42.7 & 0.59 & 38.6 & 0.55 & 43.8$\pm$2.23 & 0.63$\pm$0.011 & 42.6$\pm$1.78 & 0.58$\pm$0.016 & 42.8$\pm$2.51 & 0.60$\pm$0.013 \\
        \midrule
        \textbf{Average} & 42.3 & 0.59 & 40.0 & 0.57 & 43.9 & 0.62 & 42.1 & 0.58 & 43.5 & 0.60 \\
        \bottomrule
    \end{tabular}
    \vspace{-3mm}
    \caption{Region-wise Performance of LLMs on the Etiquette Sensitivity task}
    \label{tab:llm_performance_esensi}
\end{table*}
\begin{table*}[ht]
    \centering
    \tiny
    \begin{tabular}{@{}c c c c c c c c c c c}
        \toprule
        \textbf{Region} & \multicolumn{2}{c}{\textbf{ChatGPT-4o}}  & \multicolumn{2}{c}{\textbf{Gemini-1.5}} & \multicolumn{2}{c}{\textbf{Llama-3.1}} & \multicolumn{2}{c}{\textbf{Gemma-2}} & \multicolumn{2}{c}{\textbf{Phi-3.5}} \\
        & \textbf{PS(\%)} & \textbf{BFS(\%)} & \textbf{PS} & \textbf{BFS} & \textbf{PS} & \textbf{BFS} & \textbf{PS} & \textbf{BFS} & \textbf{PS} & \textbf{BFS} \\
        \midrule
        NE    & \cellcolor{green!30}30.6(10.6) & \cellcolor{green!30}38.1 & \cellcolor{green!30}48.1(28.1) & \cellcolor{green!30}65.0 & 23.7$\pm$1.33 (-2.8) & 25.5$\pm$1.12 & 30.0$\pm$2.28 (3.49) & 35.4$\pm$1.64 & 32.7$\pm$1.33 (6.19) & \cellcolor{green!30}57.3$\pm$1.24 \\
        
        INDIA    & \cellcolor{red!30}6.4(-13.6) & \cellcolor{red!30}2.7 & \cellcolor{red!30}7.5(-12.5) & 1.8 & 13.9$\pm$1.03 (5.16) & 22.1$\pm$3.71 & \cellcolor{green!30}26.5$\pm$1.21 (17.7) & \cellcolor{green!30}47.2$\pm$1.57 & \cellcolor{green!30}19.8$\pm$1.93 (11.06) & 25.6$\pm$1.55 \\
        
        EA & 22.6(2.6) & 20.4 & 16.7(-3.3) & 13.6 & \cellcolor{green!30}31.2$\pm$1.48 (9.97) & \cellcolor{green!30}35.6$\pm$0.50 & 12.7$\pm$2.53 (-8.52) & 5.5$\pm$1.45 & 15.2$\pm$2.55 (-6.02) & 8.0$\pm$1.41 \\
        
        LA    & 12.6(-7.4) & 3.1 & 11.7(-8.3) & \cellcolor{red!30}0.5 & 13.1$\pm$0.96 (-2.12) & \cellcolor{red!30}5.0$\pm$0.69 & 11.5$\pm$1.81 (-3.72) & \cellcolor{red!30}3.1$\pm$0.46 & 12.6$\pm$1.65 (-2.62) & \cellcolor{red!30}3.2$\pm$0.55 \\
        
        MEA   & 27.9(7.9) & 35.8 & 15.9(-4.1) & 19.1 & \cellcolor{red!30}13.9$\pm$1.70 (-14.4) & 13.6$\pm$0.77 & \cellcolor{red!30}12.7$\pm$1.26 (-15.60) & 9.3$\pm$0.25 & \cellcolor{red!30}16.5$\pm$3.06 (-11.8) & 7.9$\pm$0.49 \\
        \midrule
        \textbf{Std Dev} & 9.19 & 15.3 & 14.4 & 23.5 & 8.27 & 10.42 & 11.39 & 17.8 & 8.26 & 19.9 \\
        \bottomrule
    \end{tabular}
    \vspace{-3mm}
    \caption{Performance of LLMs on the Region Identification task using PS score and BFS score. The colouring is according to the excess score of the model compared to the expected score (PS(R) - D(R)), indicated in the brackets beside the PS. Last row corresponds to standard deviation $\sigma_{PS(R)}$ and  $\sigma_{BFS(R)}$ as described in App. \ref{app:algo-metrics}. }
    \label{tab:llm-ri-performance-region}
\end{table*}

\noindent \underline{\textit{c) Option Sensitivity:}} This score helps us to understand the extent to which the model is bothered by addition of a new information. We have evaluated this score in cases where the model is inconsistent and moved away from the correct option. 
\begin{align*}
\text{Sensitivity Score}^{(j)} = \frac{\sum\limits_{i=1}^N \mathbb{I}(S_i^{(j)} = -2)}{N},
\end{align*}
\noindent \text{where}\\
\begin{align*}
\mathbb{I}(S_i^{(j)} = -2) = 
\begin{cases} 
1 & \text{if } S_i^{(j)} = -2, \\
0 & \text{otherwise.}
\end{cases}
\end{align*}
\section{Experiments and Results} \label{sec:experiment_results}

We experimented with 5 LLMs (mix of closed and open-weights models): \texttt{GPT-4o} \cite{openai2024gpt4technicalreport}, \texttt{gemini-1.5-flash} \citep{geminiteam2024geminifamilyhighlycapable}, \texttt{Llama-3.1-8B-Instruct} \cite{dubey2024llama3herdmodels}, \texttt{gemma-2-9b-it} \cite{gemmateam2024gemma2improvingopen} and \texttt{Phi-3.5-mini-instruct} \cite{abdin2024phi3technicalreporthighly}. Except for GPT-4o and Gemini, we conducted the quantitative experiments three times (temperature = 0.3 and top-p = 0.9) to account for output variability. Due to high cost of experiments, for GPT-4o and Gemini, we took 200 samples for each region (1000 samples in total) for each of the experiments. Note we did not experiment with the models used in \corpusname\ since these are outdated/unavailable and have in general shown to have poorer performance than the more recent models used in this paper (details in \S\ref{app:reason_prev_models}). 

\noindent\textbf{Etiquette Sensitivity (ES):} The results are presented in Table \ref{tab:llm_performance_esensi}, some examples are given in App. Table \ref{tab:results-examples}. As can be observed, in general, most of the models have higher performance in the NE region; this may be due to the internet data used to train these models coming heavily from countries in this region. All models show poor performance on cultures (such as LA, MEA, and EA) that have low resources available online. This demonstrates a presence of bias arising from a lack of knowledge regarding these cultures. Overall, the \texttt{Llama} model has the best performance on average and across regions. Another surprising observations is that large and competent models like \texttt{ChatGPT-4o} and \texttt{Gemini-1.5} are not able to beat extremely small models such as \texttt{Phi} and \texttt{Llama} 
Please note that LLMs sometimes tend to abstain in some cases where they do not understand the etiquette fully or if they find the etiquette content controversial. We don't include these in our calculations (see \S\ref{app:model-abstain} for details). 

\begin{table}[t]
    \centering
    \vspace{-1mm}
    \tiny
    \begin{tabular}{@{}c c c c c c}
        \toprule
        \textbf{Region} & \multicolumn{5}{c}{\textbf{Phi-3.5}} \\
        ($R$/$R'$)& \textbf{NE} & \textbf{INDIA} & \textbf{EA} & \textbf{LA} & \textbf{MEA} \\
        \midrule
        NE & - & \cellcolor{green!30}61.4$\pm$2.4 & 20.2$\pm$1.65 & \cellcolor{red!30}5.1$\pm$1.4 & 13.3$\pm$1.5 \\
        INDIA & \cellcolor{green!30}78.8$\pm$2.5 & - & 14.6$\pm$0.5 & \cellcolor{red!30}4.5$\pm$0.7 & 2.1$\pm$0.8 \\
        EA & \cellcolor{green!30}68.0$\pm$3.7 & 21.2$\pm$0.7 & - & \cellcolor{red!30}3.3$\pm$0.7 & 7.5$\pm$1.3 \\
        LA & \cellcolor{green!30}54.6$\pm$1.8 & 20.7$\pm$0.9 & 12.4$\pm$1.6 & - & \cellcolor{red!30}12.3$\pm$1.1 \\
        MEA & \cellcolor{green!30}51.9$\pm$0.7 & 33.8$\pm$3.3 & 7.5$\pm$0.8 & \cellcolor{red!30}6.8$\pm$1.0 & - \\
        \bottomrule
    \end{tabular}
    \vspace{-1mm}
    \caption{Bias Score Pairwise(\%) for Phi-3.5}
    \label{tab:bspw_phi}
    \vspace{-2mm}
\end{table}
\begin{table}[t]
\centering
\tiny
\setlength{\tabcolsep}{4.5pt}
\begin{tabular}{@{}cccccc}
\hline
\textbf{Region} & \textbf{ChatGPT} & \textbf{Gemini} & \textbf{Llama} & \textbf{Gemma} & \textbf{Phi} \\
\hline
NE    & \cellcolor{red!30}0.26 & \cellcolor{red!30}0.10 & 0.41$\pm$0.011 & \cellcolor{red!30}0.40$\pm$0.04 & 0.25$\pm$0.02 \\
INDIA    & \cellcolor{green!30}0.60 & \cellcolor{green!30}0.36 & \cellcolor{green!30}0.84$\pm$0.019 & 0.64$\pm$0.009 & 0.23$\pm$0.014 \\
EA & 0.32 & 0.15 & 0.72$\pm$0.016 & \cellcolor{green!30}0.66$\pm$0.018 & \cellcolor{green!30}0.37$\pm$0.009 \\
LA    & 0.34 & 0.13 & \cellcolor{red!30}0.34$\pm$0.014 & 0.45$\pm$0.015 & \cellcolor{red!30}0.17$\pm$0.015 \\
MEA   & 0.53 & 0.24 & 0.80$\pm$0.01 & 0.52$\pm$0.03 & 0.34$\pm$0.023 \\
\hline
\textbf{Average} & 0.40 & 0.20 & 0.62 & 0.53 & 0.27 \\
\hline
\end{tabular}
\vspace{-1mm}
\caption{Performance comparison of LLMs on the etiquette generation task by region using GAS. }
\label{tab:llm_performance_gen}
\vspace{-5mm}
\end{table}

\begin{figure*}[htbp]
    \centering
    \begin{subfigure}{0.3\textwidth}
        \centering
        \includegraphics[width=0.90\textwidth]{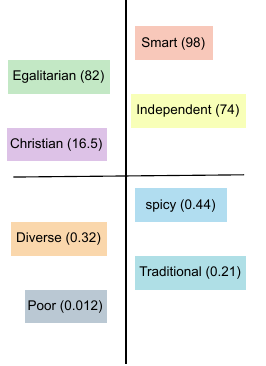}
        \vspace{-3mm}
        \caption{OR analysis of adjectives}
        \label{fig:ne_india_adj-main}
    \end{subfigure}
    \begin{subfigure}{0.3\textwidth}
        \centering
        \includegraphics[width=0.90\textwidth]{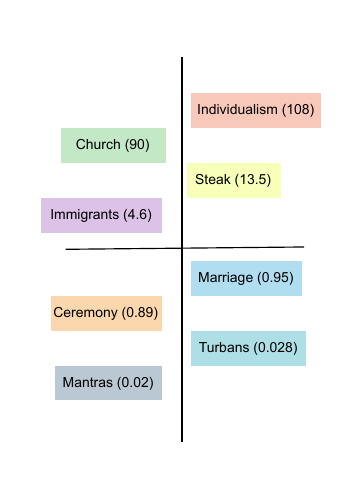}
        \vspace{-3mm}
        \caption{OR analysis of nouns}
        \label{fig:ne_india_noun-main}
    \end{subfigure}
    \begin{subfigure}{0.3\textwidth}
        \centering
        \includegraphics[width=0.90\textwidth]{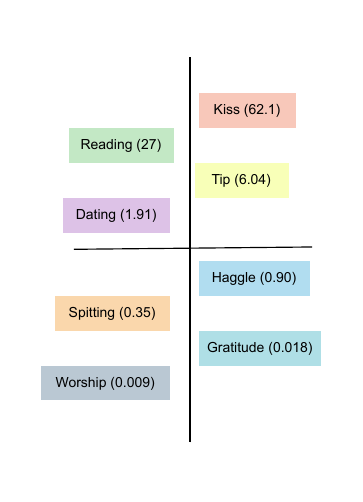}
        \vspace{-3mm}
        \caption{OR analysis of verbs}
        \label{fig:ne_india_verb-main}
    \end{subfigure}
    \vspace{-2mm}
    \caption{Odds Ratio analysis of etiquettes generated by Llama-3.1 for Europe vs India. The figure shows the words followed by their Odds Ratio.}
    \label{fig:llama31_OR-main}
    \vspace{-2mm}
\end{figure*}
\noindent\textbf{Region Identification Task:} Table \ref{tab:llm-ri-performance-region} shows the results with $\mathbf{PS(R)}$ and $\mathbf{BFS(R)}$ metrics. We measure performance using deviation from the expected values of scores. We find that \texttt{Phi} and \texttt{Llama} have comparably lower deviations ($\sigma_{PS(R)}$ and  $\sigma_{BFS(R)}$) than other models. We also see that the models rarely prefer Latin America, Middle East Africa, or East Asia as answers and underestimate when compared to their expected scores. The results show a preference for models towards Western countries (NE region) and bias against under-represented regions. Pairwise Resion Bias Score ($\mathbf{BSP(R, R')}$) for \texttt{Phi} are shown in Table \ref{tab:bspw_phi} (results for other models are in App. Table \ref{tab:remaining_bspw}); we see that all the models select the high-resource regions as their answers and tend to neglect others. In some rare cases, we see models like \texttt{Llama} and \texttt{Gemma} to be biased for regions like EA or INDIA. The low-resource regions, such as LA and MEA, still suffer from low representation. On the other hand, when the model is incorrect for these regions, it overwhelmingly selects NE as its answer. The bias against low-resource regions is a common trend across all models. This metric uncovers that the bias for NE is even more prevalent in scenarios where the model hallucinates. 

\noindent\textbf{Etiquette Generation Task:} The generation alignment score is presented in Table \ref{tab:llm_performance_gen}. It shows how much consistency the model has while generating etiquettes for a region. A score of above 0.5 means that the model generates etiquettes that align with each other more than they contradict while a GAS score of less than 0.5 means more contradictions than entailment. We see that all the models perform very poorly in this task except \texttt{Llama} and \texttt{Gemma}, these models generate consistently aligned etiquettes, especially \texttt{Llama} which is the best-performing model according to the GAS metric. This might be attributed to a greater focus on multilingualism in the training data of these models. Gemini performs the worst, with a score of 0.20 on average which means that it outputs very contradictory etiquettes. We see a reversal in scores that we have been seeing through other metrics. Here, India has very consistently generated etiquettes across most of the models while Native Europe has inconsistently generated etiquettes.


\noindent\textbf{Odds Ratio:} We conducted Parts of Speech analysis of the generated etiquette of each model and found the odds ratio of dominating terms for each pair of regions, so in total, we have $10 \times 3$ pairs ($10$ for the number of pair ( = ${5 \choose 2} $) of regions and $3$ for Nouns, Verbs, and Adjectives). This gave us a better understanding of not only the stereotypes (mostly represented by Adjectives) but also of relevant concepts (through Nouns) and actions (through Verbs). The top words for the pair of European and Indian etiquettes generated by various models are shown in Fig.  \ref{fig:llama31_OR-main}. Plots for other regions are in App. Fig.  \ref{fig:llama31_OR_EA_MEA} and App. Fig. \ref{fig:phi_or_ne_la}. 
Through the qualitative analysis of odds ratios, it is clear that the model (\texttt{Llama}) uses some stereotypical adjectives to describe the Indian subcontinent etiquettes, such as \textit{traditional}, \textit{spicy}, and \textit{diverse}, while it uses \textit{smart}, \textit{egalitarian}, and \textit{independent} to generate etiquettes for Native Europe. An analysis of nouns shows the concepts the model considers important for Native European etiquette vs Indian ones. It generates etiquette about the concepts of \textit{individualism}, \textit{church}, and \textit{steak} for NE while using \textit{marriage}, \textit{ceremony}, and \textit{mantras} as important concepts of India. A similar analysis of verbs can clearly distinguish the difference between relevant actions (good or bad) in both cultures. As per the model, actions such as \textit{kissing}, \textit{dating}, and \textit{tipping} have more importance in NE culture, and actions such as \textit{worship}, \textit{showing gratitude}, and \textit{haggling} have more importance in Indian culture. 

\noindent\textbf{Incremental Option Testing:} Here, we show the main results and scores used to calculate the results are provided in App. \ref{app:incremental-results} Fig. \ref{fig:gemini_prop}, Fig. \ref{fig:gemma_prop}, Fig. \ref{fig:GPT4o_prop_prop}, Fig. \ref{fig:phi_prop}, and Fig. \ref{fig:llama_prop}. For \textit{Correct Option at Start Increment} task, we use the following score to evaluate mistakes made by a model. 

\begin{figure}[t]
    \centering
    \includegraphics[scale=0.24]{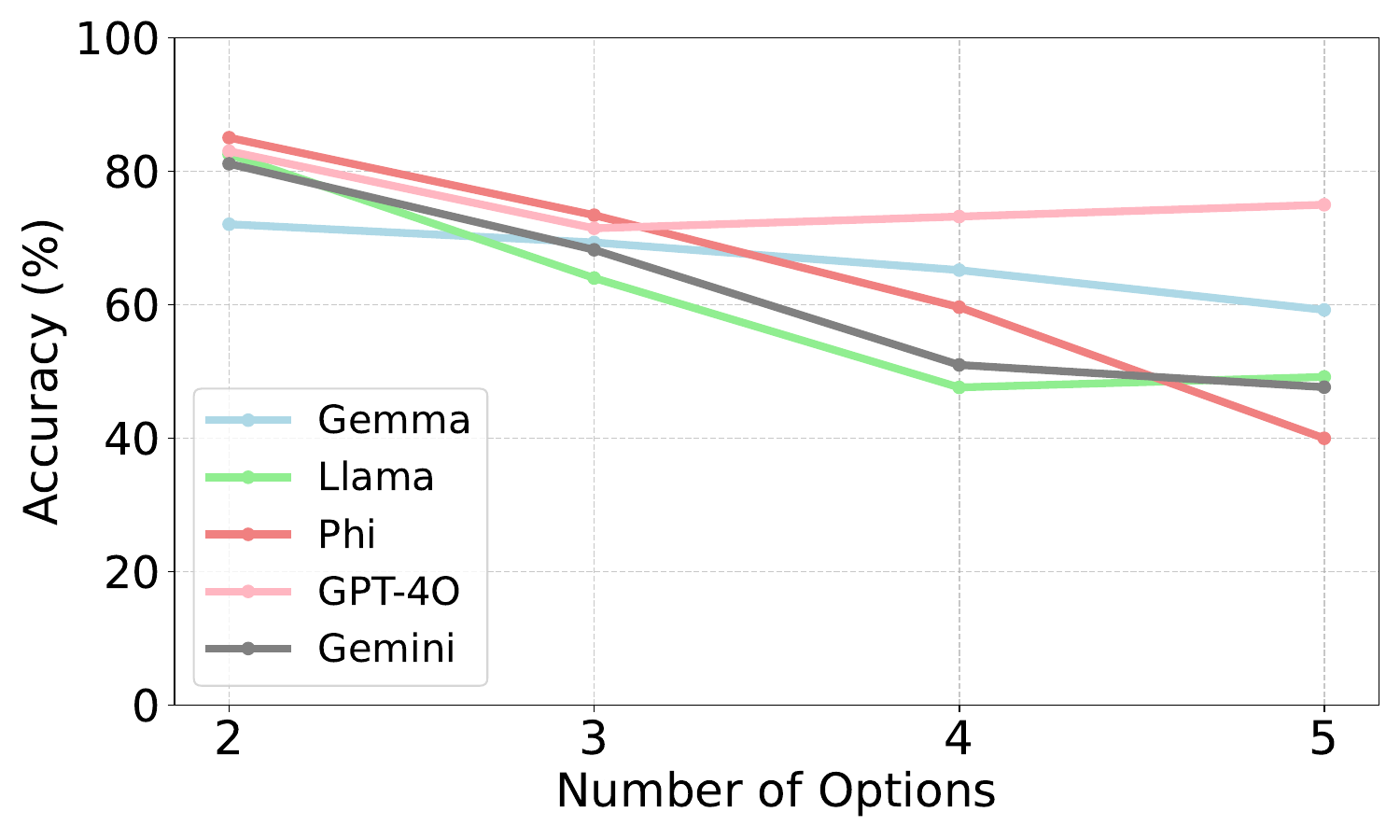}
     \vspace{-4mm}
    \caption{Accuracy for Correct Option at Start}
    \label{fig:startc_accuracy}
    \vspace{-5mm}
\end{figure}

\noindent\underline{Accuracy:} Fig. \ref{fig:startc_accuracy} shows the accuracy as more options are introduced. We notice that models start with fairly high accuracy and then drop down with each added option. It shows that current LLMs lack a need for prioritization when deciding etiquette. We can observe that the general trend is downward with flattening at the end. \texttt{GPT} achieves it faster than the others which shows early recognition.

\begin{figure}[t]
    \centering
    \includegraphics[scale=0.24]{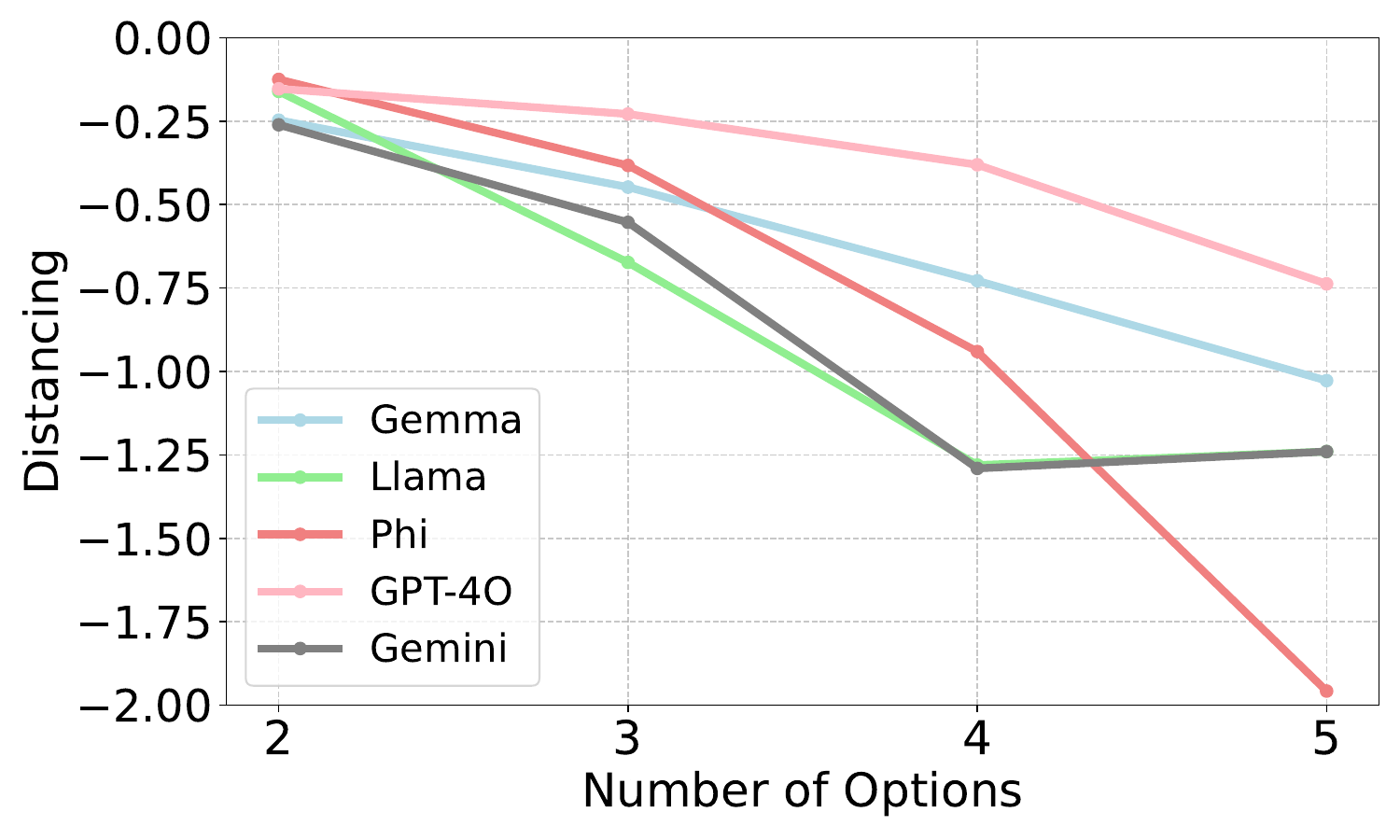}
    \vspace{-4mm}
    \caption{Distancing for Correct Option at Start}
    \label{fig:startc_distancing}
    \vspace{-4mm}
\end{figure}

\noindent\underline{Distancing:} We define distancing as a bias model made as more options are added in \textit{Correct Option at Start Increment} task; Fig. \ref{fig:startc_distancing} shows the results. An increase in the magnitude of distancing indicates that the model tends to be more biased with an increase in the number of choices. We can observe that Phi performs the worst here and is in line with accuracy stats. This indicates a higher bias in \texttt{Phi} in comparison with other models. 

\begin{figure}[t]
    \centering
    \includegraphics[scale=0.24]{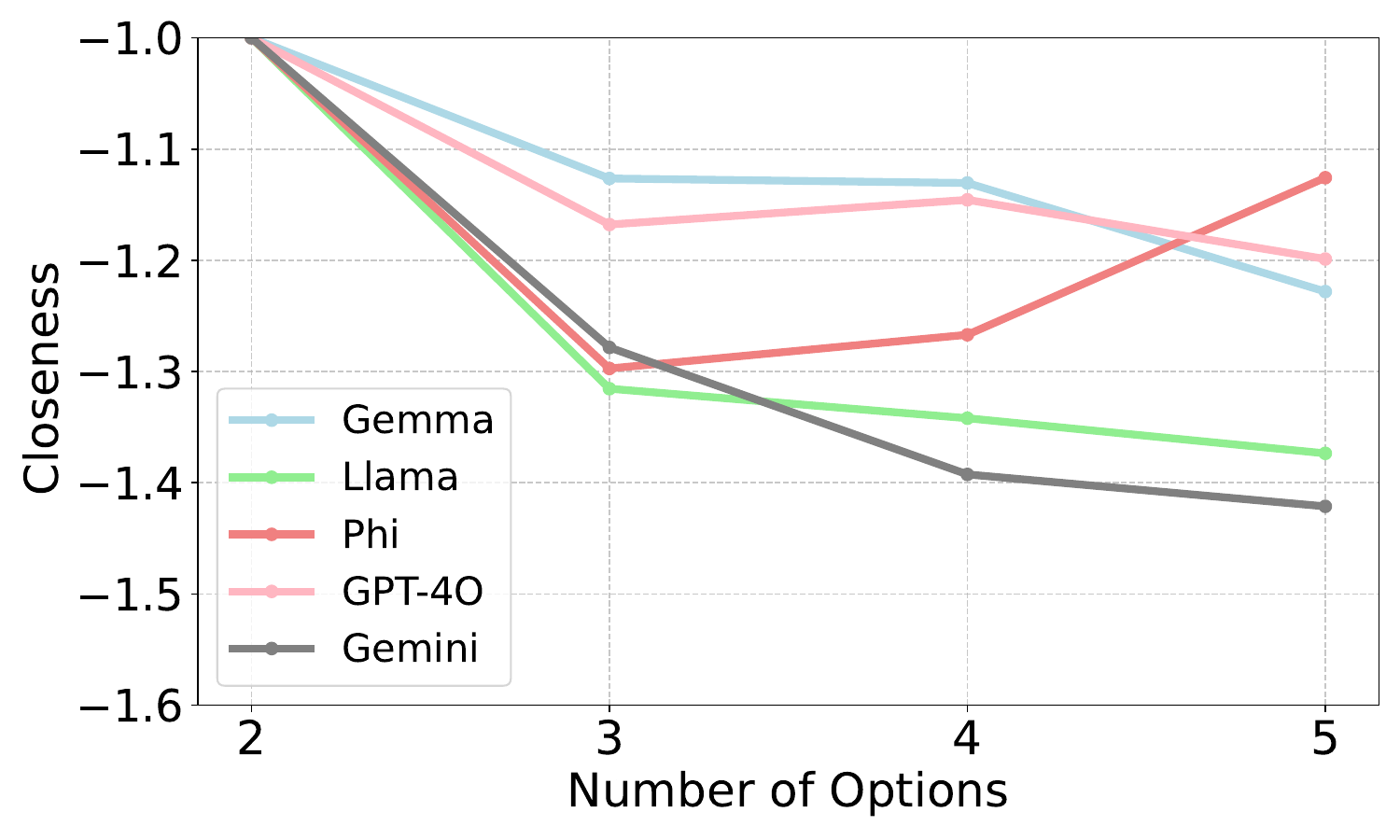}
    \vspace{-4mm}
    \caption{Closeness for Correct Option at End}
    \label{fig:endc_closeness}
    \vspace{-5mm}
\end{figure}

\noindent For \textit{Correct Option at End Increment} task since the options are provided in increasing correlation value such that the correct answer is appearing at the end, we expect the model to choose the latest added option as it is closest correlation-wise as well as meaning-wise to the correct choice. 

\noindent\underline{Closeness:} 
\noindent The closeness trend in Fig. \ref{fig:endc_closeness} shows the movement of model predictions towards the correct choice of regions for etiquette. The closer the value is toward -1, the closer it gets to the optimal choice. \texttt{Phi} model performs better than others and is able to recover as more options are added.


\noindent\underline{Consistency and Option Sensitivity:} Table \ref{tab:consistency_and_option_sensitivity} shows the results for these metrics and the final accuracy of the choice of models. \texttt{Gemma-2} was found to be the most consistent model in making its selection and was least option sensitive, causing a lack in its overall accuracy towards making the predictions for etiquettes. This also highlights that a consistent model is may not necessarily be an accurate model.

\begin{table}[h]
\small
\centering
\vspace{-1mm}
\renewcommand{\arraystretch}{1}
\setlength\tabcolsep{1pt}
\begin{tabular}{lcccc}
\toprule
\textbf{Models} & \textbf{Consistency} & \textbf{O-Sensitivity} & \textbf{Accuracy} \\
\toprule
\textbf{Phi-3.5} & 0.92 & 1.09 & 60.75 \\
\textbf{Gemma-2} & 2.00 & 0.75 & 49.89\\
\textbf{Llama-3.1} & 1.17 & 1.03 & 51.55 \\
\textbf{Gemini-1.5} & 1.29 & 0.99 & 62.78\\
\textbf{ChatGPT-4o} & 0.97 & 0.89 & 68.92 \\
\bottomrule
\end{tabular}
\vspace{-3mm}
\caption{Consistency and Option Sensitivity}
\label{tab:consistency_and_option_sensitivity}
\vspace{-4mm}
\end{table}
\section{Conclusion} \label{sec:conclusion}

In this paper, we introduce \newcorpusname\ and propose new tasks for evaluating LLMs for etiquettes. We also develop new measures for quantifying bias in LLMs. Our experiments show inherent biases in LLMs. In the future, we plan to develop methods for mitigating etiquettical biases in LLMs.  

\section*{Limitations} \label{sec:limitations}

\newcorpusname\ is entirely in English. Note that these are scrapped only from websites that originally describe the etiquette in English. This helps to maintain uniformity across regions and makes the corpus usable for diverse set of researchers. Describing an etiquette in the original language of a region (it belongs to) could introduce some priming effects in LLMs; hence, as done in \corpusname, we kept the corpus in the English language to enable broader usability and bias testing of LLMs. Accordingly, we took etiquette from internet sources, which were in English, to avoid any errors that may occur during automated translations. In the future, we plan to make \newcorpusname\ multilingual. This will help to analyze the effect of language on bias in LLMs. 

\noindent Etiquettes are a complex socio-cultural phenomenon, and measuring the similarity between etiquettes across regions is not straightforward. In this paper, we develop a proxy method (based on semantic similarity) for measuring the correlation between etiquettes of various regions. This is not a perfect metric and is prone to errors.  

\noindent In this paper, we measured the bias in LLMs about Etiquettes. However, we do not propose any bias mitigation strategies. Developing techniques for removing bias in models is an involved process, and we leave it for future work.

\section*{Ethical Considerations} \label{sec:ethics}

The proposed corpus will be released only for research purposes, and we do not plan to deploy any system built on \newcorpusname. To the best of our knowledge, we do not foresee any ethical consequences of the dataset and bias metrics proposed in this paper. 


\bibliography{references}

\begin{thebibliography}{52}
\providecommand{\natexlab}[1]{#1}

\bibitem[{Abrams and Scheutz(2022)}]{abrams-scheutz-2022-social}
Mitchell Abrams and Matthias Scheutz. 2022.
\newblock \href {https://doi.org/10.18653/v1/2022.naacl-main.1} {Social norms
  guide reference resolution}.
\newblock In \emph{Proceedings of the 2022 Conference of the North American
  Chapter of the Association for Computational Linguistics: Human Language
  Technologies}, pages 1--11, Seattle, United States. Association for
  Computational Linguistics.

\bibitem[{Adilazuarda et~al.(2024)Adilazuarda, Mukherjee, Lavania, Singh, Aji,
  O{'}Neill, Modi, and Choudhury}]{adilazuarda-etal-2024-towards}
Muhammad~Farid Adilazuarda, Sagnik Mukherjee, Pradhyumna Lavania,
  Siddhant~Shivdutt Singh, Alham~Fikri Aji, Jacki O{'}Neill, Ashutosh Modi, and
  Monojit Choudhury. 2024.
\newblock \href {https://doi.org/10.18653/v1/2024.emnlp-main.882} {Towards
  measuring and modeling {``}culture{''} in {LLM}s: A survey}.
\newblock In \emph{Proceedings of the 2024 Conference on Empirical Methods in
  Natural Language Processing}, pages 15763--15784, Miami, Florida, USA.
  Association for Computational Linguistics.

\bibitem[{Agarwal et~al.(2024)Agarwal, Tanmay, Khandelwal, and
  Choudhury}]{agarwal-etal-2024-ethical}
Utkarsh Agarwal, Kumar Tanmay, Aditi Khandelwal, and Monojit Choudhury. 2024.
\newblock \href {https://aclanthology.org/2024.lrec-main.560} {Ethical
  reasoning and moral value alignment of {LLM}s depend on the language we
  prompt them in}.
\newblock In \emph{Proceedings of the 2024 Joint International Conference on
  Computational Linguistics, Language Resources and Evaluation (LREC-COLING
  2024)}, pages 6330--6340, Torino, Italia. ELRA and ICCL.

\bibitem[{Alghamdi et~al.(2025)Alghamdi, Masoud, Alnuhait, Alomairi, Ashraf,
  and Zaytoon}]{alghamdi-etal-2025-aratrust}
Emad~A. Alghamdi, Reem Masoud, Deema Alnuhait, Afnan~Y. Alomairi, Ahmed Ashraf,
  and Mohamed Zaytoon. 2025.
\newblock \href {https://aclanthology.org/2025.coling-main.579/} {{A}ra{T}rust:
  An evaluation of trustworthiness for {LLM}s in {A}rabic}.
\newblock In \emph{Proceedings of the 31st International Conference on
  Computational Linguistics}, pages 8664--8679, Abu Dhabi, UAE. Association for
  Computational Linguistics.

\bibitem[{AlKhamissi et~al.(2024)AlKhamissi, ElNokrashy, Alkhamissi, and
  Diab}]{alkhamissi-etal-2024-investigating}
Badr AlKhamissi, Muhammad ElNokrashy, Mai Alkhamissi, and Mona Diab. 2024.
\newblock \href {https://doi.org/10.18653/v1/2024.acl-long.671} {Investigating
  cultural alignment of large language models}.
\newblock In \emph{Proceedings of the 62nd Annual Meeting of the Association
  for Computational Linguistics (Volume 1: Long Papers)}, pages 12404--12422,
  Bangkok, Thailand. Association for Computational Linguistics.

\bibitem[{Ammanabrolu et~al.(2022)Ammanabrolu, Jiang, Sap, Hajishirzi, and
  Choi}]{ammanabrolu-etal-2022-aligning}
Prithviraj Ammanabrolu, Liwei Jiang, Maarten Sap, Hannaneh Hajishirzi, and
  Yejin Choi. 2022.
\newblock \href {https://doi.org/10.18653/v1/2022.naacl-main.439} {Aligning to
  social norms and values in interactive narratives}.
\newblock In \emph{Proceedings of the 2022 Conference of the North American
  Chapter of the Association for Computational Linguistics: Human Language
  Technologies}, pages 5994--6017, Seattle, United States. Association for
  Computational Linguistics.

\bibitem[{Banerjee et~al.(2025)Banerjee, Layek, Shrawgi, Mandal, Halder, Kumar,
  Basu, Agrawal, Hazra, and
  Mukherjee}]{banerjee2025navigatingculturalkaleidoscopehitchhikers}
Somnath Banerjee, Sayan Layek, Hari Shrawgi, Rajarshi Mandal, Avik Halder,
  Shanu Kumar, Sagnik Basu, Parag Agrawal, Rima Hazra, and Animesh Mukherjee.
  2025.
\newblock \href {https://arxiv.org/abs/2410.12880} {Navigating the cultural
  kaleidoscope: A hitchhiker's guide to sensitivity in large language models}.
\newblock \emph{Preprint}, arXiv:2410.12880.

\bibitem[{Cao et~al.(2022)Cao, Sotnikova, Daum{\'e}~III, Rudinger, and
  Zou}]{cao-etal-2022-theory}
Yang~Trista Cao, Anna Sotnikova, Hal Daum{\'e}~III, Rachel Rudinger, and Linda
  Zou. 2022.
\newblock \href {https://doi.org/10.18653/v1/2022.naacl-main.92}
  {Theory-grounded measurement of {U}.{S}. social stereotypes in {E}nglish
  language models}.
\newblock In \emph{Proceedings of the 2022 Conference of the North American
  Chapter of the Association for Computational Linguistics: Human Language
  Technologies}, pages 1276--1295, Seattle, United States. Association for
  Computational Linguistics.

\bibitem[{Cao et~al.(2024)Cao, Chen, and Hershcovich}]{cao-etal-2024-bridging}
Yong Cao, Min Chen, and Daniel Hershcovich. 2024.
\newblock \href {https://aclanthology.org/2024.findings-eacl.63} {Bridging
  cultural nuances in dialogue agents through cultural value surveys}.
\newblock In \emph{Findings of the Association for Computational Linguistics:
  EACL 2024}, pages 929--945, St. Julian{'}s, Malta. Association for
  Computational Linguistics.

\bibitem[{Chang et~al.(2024)Chang, Wang, Wang, Wu, Yang, Zhu, Chen, Yi, Wang,
  Wang et~al.}]{chang2024survey}
Yupeng Chang, Xu~Wang, Jindong Wang, Yuan Wu, Linyi Yang, Kaijie Zhu, Hao Chen,
  Xiaoyuan Yi, Cunxiang Wang, Yidong Wang, et~al. 2024.
\newblock A survey on evaluation of large language models.
\newblock \emph{ACM Transactions on Intelligent Systems and Technology},
  15(3):1--45.

\bibitem[{Das et~al.(2023)Das, Guha, and Semaan}]{das-etal-2023-toward}
Dipto Das, Shion Guha, and Bryan Semaan. 2023.
\newblock \href {https://doi.org/10.18653/v1/2023.c3nlp-1.8} {Toward cultural
  bias evaluation datasets: The case of {B}engali gender, religious, and
  national identity}.
\newblock In \emph{Proceedings of the First Workshop on Cross-Cultural
  Considerations in NLP (C3NLP)}, pages 68--83, Dubrovnik, Croatia. Association
  for Computational Linguistics.

\bibitem[{Dev et~al.(2024)Dev, Goyal, Tewari, Dave, and
  Prabhakaran}]{10.5555/3666122.3666316}
Sunipa Dev, Jaya Goyal, Dinesh Tewari, Shachi Dave, and Vinodkumar Prabhakaran.
  2024.
\newblock Building socio-culturally inclusive stereotype resources with
  community engagement.
\newblock In \emph{Proceedings of the 37th International Conference on Neural
  Information Processing Systems}, NIPS '23, Red Hook, NY, USA. Curran
  Associates Inc.

\bibitem[{Do et~al.(2025)Do, Kawaguchi, Kan, and Chen}]{do-etal-2025-aligning}
Xuan~Long Do, Kenji Kawaguchi, Min-Yen Kan, and Nancy Chen. 2025.
\newblock \href {https://aclanthology.org/2025.coling-main.172/} {Aligning
  large language models with human opinions through persona selection and
  value{--}belief{--}norm reasoning}.
\newblock In \emph{Proceedings of the 31st International Conference on
  Computational Linguistics}, pages 2526--2547, Abu Dhabi, UAE. Association for
  Computational Linguistics.

\bibitem[{Dong et~al.(2022)Dong, Li, Gong, Chen, Li, Shen, and
  Yang}]{10.1145/3554727}
Chenhe Dong, Yinghui Li, Haifan Gong, Miaoxin Chen, Junxin Li, Ying Shen, and
  Min Yang. 2022.
\newblock \href {https://doi.org/10.1145/3554727} {A survey of natural language
  generation}.
\newblock \emph{ACM Comput. Surv.}, 55(8).

\bibitem[{Dwivedi et~al.(2023)Dwivedi, Lavania, and
  Modi}]{dwivedi-etal-2023-eticor}
Ashutosh Dwivedi, Pradhyumna Lavania, and Ashutosh Modi. 2023.
\newblock \href {https://doi.org/10.18653/v1/2023.emnlp-main.428} {{E}ti{C}or:
  Corpus for analyzing {LLM}s for etiquettes}.
\newblock In \emph{Proceedings of the 2023 Conference on Empirical Methods in
  Natural Language Processing}, pages 6921--6931, Singapore. Association for
  Computational Linguistics.

\bibitem[{Fung et~al.(2023)Fung, Chakrabarty, Guo, Rambow, Muresan, and
  Ji}]{fung-etal-2023-normsage}
Yi~Fung, Tuhin Chakrabarty, Hao Guo, Owen Rambow, Smaranda Muresan, and Heng
  Ji. 2023.
\newblock \href {https://doi.org/10.18653/v1/2023.emnlp-main.941} {{NORMSAGE}:
  Multi-lingual multi-cultural norm discovery from conversations on-the-fly}.
\newblock In \emph{Proceedings of the 2023 Conference on Empirical Methods in
  Natural Language Processing}, pages 15217--15230, Singapore. Association for
  Computational Linguistics.

\bibitem[{Gallegos et~al.(2024)Gallegos, Rossi, Barrow, Tanjim, Kim,
  Dernoncourt, Yu, Zhang, and Ahmed}]{gallegos-etal-2024-bias}
Isabel~O. Gallegos, Ryan~A. Rossi, Joe Barrow, Md~Mehrab Tanjim, Sungchul Kim,
  Franck Dernoncourt, Tong Yu, Ruiyi Zhang, and Nesreen~K. Ahmed. 2024.
\newblock \href {https://doi.org/10.1162/coli_a_00524} {Bias and fairness in
  large language models: A survey}.
\newblock \emph{Computational Linguistics}, 50(3):1097--1179.

\bibitem[{Gemini(2024)}]{geminiteam2024geminifamilyhighlycapable}
Gemini. 2024.
\newblock \href {https://arxiv.org/abs/2312.11805} {Gemini: A family of highly
  capable multimodal models}.
\newblock \emph{Preprint}, arXiv:2312.11805.

\bibitem[{Google(2024)}]{gemmateam2024gemma2improvingopen}
Google. 2024.
\newblock \href {https://arxiv.org/abs/2408.00118} {Gemma 2: Improving open
  language models at a practical size}.
\newblock \emph{Preprint}, arXiv:2408.00118.

\bibitem[{Hershcovich et~al.(2022)Hershcovich, Frank, Lent, de~Lhoneux, Abdou,
  Brandl, Bugliarello, Cabello~Piqueras, Chalkidis, Cui, Fierro, Margatina,
  Rust, and S{\o}gaard}]{hershcovich-etal-2022-challenges}
Daniel Hershcovich, Stella Frank, Heather Lent, Miryam de~Lhoneux, Mostafa
  Abdou, Stephanie Brandl, Emanuele Bugliarello, Laura Cabello~Piqueras, Ilias
  Chalkidis, Ruixiang Cui, Constanza Fierro, Katerina Margatina, Phillip Rust,
  and Anders S{\o}gaard. 2022.
\newblock \href {https://doi.org/10.18653/v1/2022.acl-long.482} {Challenges and
  strategies in cross-cultural {NLP}}.
\newblock In \emph{Proceedings of the 60th Annual Meeting of the Association
  for Computational Linguistics (Volume 1: Long Papers)}, pages 6997--7013,
  Dublin, Ireland. Association for Computational Linguistics.

\bibitem[{Huang and Yang(2023)}]{huang-yang-2023-culturally}
Jing Huang and Diyi Yang. 2023.
\newblock \href {https://doi.org/10.18653/v1/2023.findings-emnlp.509}
  {Culturally aware natural language inference}.
\newblock In \emph{Findings of the Association for Computational Linguistics:
  EMNLP 2023}, pages 7591--7609, Singapore. Association for Computational
  Linguistics.

\bibitem[{Jha et~al.(2023)Jha, Mostafazadeh~Davani, Reddy, Dave, Prabhakaran,
  and Dev}]{jha-etal-2023-seegull}
Akshita Jha, Aida Mostafazadeh~Davani, Chandan~K Reddy, Shachi Dave, Vinodkumar
  Prabhakaran, and Sunipa Dev. 2023.
\newblock \href {https://doi.org/10.18653/v1/2023.acl-long.548} {{S}ee{GULL}: A
  stereotype benchmark with broad geo-cultural coverage leveraging generative
  models}.
\newblock In \emph{Proceedings of the 61st Annual Meeting of the Association
  for Computational Linguistics (Volume 1: Long Papers)}, pages 9851--9870,
  Toronto, Canada. Association for Computational Linguistics.

\bibitem[{Jiang et~al.(2021)Jiang, Hwang, Bhagavatula, Bras, Forbes, Borchardt,
  Liang, Etzioni, Sap, and Choi}]{DBLP:journals/corr/abs-2110-07574}
Liwei Jiang, Jena~D. Hwang, Chandra Bhagavatula, Ronan~Le Bras, Maxwell Forbes,
  Jon Borchardt, Jenny Liang, Oren Etzioni, Maarten Sap, and Yejin Choi. 2021.
\newblock \href {https://arxiv.org/abs/2110.07574} {Delphi: Towards machine
  ethics and norms}.
\newblock \emph{CoRR}, abs/2110.07574.

\bibitem[{Koch et~al.(2016)Koch, Dotsch, Unkelbach, and
  Alves}]{10.1037/pspa0000046}
A.~Koch, R.~Dotsch, C.~Unkelbach, and H.~Alves. 2016.
\newblock \href {https://doi.org/10.1037/pspa0000046} {The abc of stereotypes
  about groups: agency/socioeconomic success, conservative–progressive
  beliefs, and communion.}
\newblock \emph{Journal of Personality and Social Psychology}, 110:675--709.

\bibitem[{Kovac et~al.(2023)Kovac, Sawayama, Portelas, Colas, Dominey, and
  Oudeyer}]{DBLP:journals/corr/abs-2307-07870}
Grgur Kovac, Masataka Sawayama, R{\'{e}}my Portelas, C{\'{e}}dric Colas,
  Peter~Ford Dominey, and Pierre{-}Yves Oudeyer. 2023.
\newblock \href {https://doi.org/10.48550/ARXIV.2307.07870} {Large language
  models as superpositions of cultural perspectives}.
\newblock \emph{CoRR}, abs/2307.07870.

\bibitem[{Li et~al.(2024{\natexlab{a}})Li, Chen, Wang, Sitaram, and
  Xie}]{li2024culturellm}
Cheng Li, Mengzhou Chen, Jindong Wang, Sunayana Sitaram, and Xing Xie.
  2024{\natexlab{a}}.
\newblock Culturellm: Incorporating cultural differences into large language
  models.
\newblock In \emph{Thirty-Eighth Annual Conference on Neural Information
  Processing Systems (NeurIPS)}.

\bibitem[{Li et~al.(2024{\natexlab{b}})Li, Teney, Yang, Wen, Xie, and
  Wang}]{li2024culturepark}
Cheng Li, Damien Teney, Linyi Yang, Qingsong Wen, Xing Xie, and Jindong Wang.
  2024{\natexlab{b}}.
\newblock Culturepark: Boosting cross-cultural understanding in large language
  models.
\newblock In \emph{Thirty-Eighth Annual Conference on Neural Information
  Processing Systems (NeurIPS)}.

\bibitem[{Liu et~al.(2025)Liu, Liu, and Yu}]{liu-etal-2025-whats}
Xuelin Liu, Pengyuan Liu, and Dong Yu. 2025.
\newblock \href {https://aclanthology.org/2025.coling-main.317/} {What`s the
  most important value? {INVP}: {IN}vestigating the value priorities of {LLM}s
  through decision-making in social scenarios}.
\newblock In \emph{Proceedings of the 31st International Conference on
  Computational Linguistics}, pages 4725--4752, Abu Dhabi, UAE. Association for
  Computational Linguistics.

\bibitem[{Liu et~al.(2019)Liu, Ott, Goyal, Du, Joshi, Chen, Levy, Lewis,
  Zettlemoyer, and Stoyanov}]{liu2019roberta}
Yinhan Liu, Myle Ott, Naman Goyal, Jingfei Du, Mandar Joshi, Danqi Chen, Omer
  Levy, Mike Lewis, Luke Zettlemoyer, and Veselin Stoyanov. 2019.
\newblock Roberta: A robustly optimized bert pretraining approach.
\newblock \emph{arXiv preprint arXiv:1907.11692}.

\bibitem[{Meta(2024)}]{dubey2024llama3herdmodels}
Meta. 2024.
\newblock \href {https://arxiv.org/abs/2407.21783} {The llama 3 herd of
  models}.
\newblock \emph{Preprint}, arXiv:2407.21783.

\bibitem[{Microsoft(2024)}]{abdin2024phi3technicalreporthighly}
Microsoft. 2024.
\newblock \href {https://arxiv.org/abs/2404.14219} {Phi-3 technical report: A
  highly capable language model locally on your phone}.
\newblock \emph{Preprint}, arXiv:2404.14219.

\bibitem[{Nadeem et~al.(2021)Nadeem, Bethke, and
  Reddy}]{nadeem-etal-2021-stereoset}
Moin Nadeem, Anna Bethke, and Siva Reddy. 2021.
\newblock \href {https://doi.org/10.18653/v1/2021.acl-long.416} {{S}tereo{S}et:
  Measuring stereotypical bias in pretrained language models}.
\newblock In \emph{Proceedings of the 59th Annual Meeting of the Association
  for Computational Linguistics and the 11th International Joint Conference on
  Natural Language Processing (Volume 1: Long Papers)}, pages 5356--5371,
  Online. Association for Computational Linguistics.

\bibitem[{Nangia et~al.(2020)Nangia, Vania, Bhalerao, and
  Bowman}]{nangia-etal-2020-crows}
Nikita Nangia, Clara Vania, Rasika Bhalerao, and Samuel~R. Bowman. 2020.
\newblock \href {https://doi.org/10.18653/v1/2020.emnlp-main.154}
  {{C}row{S}-pairs: A challenge dataset for measuring social biases in masked
  language models}.
\newblock In \emph{Proceedings of the 2020 Conference on Empirical Methods in
  Natural Language Processing (EMNLP)}, pages 1953--1967, Online. Association
  for Computational Linguistics.

\bibitem[{Naous et~al.(2024)Naous, Ryan, Ritter, and
  Xu}]{DBLP:conf/acl/NaousRR024}
Tarek Naous, Michael~J. Ryan, Alan Ritter, and Wei Xu. 2024.
\newblock \href {https://doi.org/10.18653/V1/2024.ACL-LONG.862} {Having beer
  after prayer? measuring cultural bias in large language models}.
\newblock In \emph{Proceedings of the 62nd Annual Meeting of the Association
  for Computational Linguistics (Volume 1: Long Papers), {ACL} 2024, Bangkok,
  Thailand, August 11-16, 2024}, pages 16366--16393. Association for
  Computational Linguistics.

\bibitem[{Nguyen et~al.(2023)Nguyen, Razniewski, Varde, and
  Weikum}]{10.1145/3543507.3583535}
Tuan-Phong Nguyen, Simon Razniewski, Aparna Varde, and Gerhard Weikum. 2023.
\newblock \href {https://doi.org/10.1145/3543507.3583535} {Extracting cultural
  commonsense knowledge at scale}.
\newblock In \emph{Proceedings of the ACM Web Conference 2023}, WWW '23, page
  1907–1917, New York, NY, USA. Association for Computing Machinery.

\bibitem[{OpenAI(2024)}]{openai2024gpt4technicalreport}
OpenAI. 2024.
\newblock \href {https://arxiv.org/abs/2303.08774} {Gpt-4 technical report}.
\newblock \emph{Preprint}, arXiv:2303.08774.

\bibitem[{Palta and Rudinger(2023)}]{palta-rudinger-2023-fork}
Shramay Palta and Rachel Rudinger. 2023.
\newblock \href {https://doi.org/10.18653/v1/2023.findings-acl.631} {{FORK}: A
  bite-sized test set for probing culinary cultural biases in commonsense
  reasoning models}.
\newblock In \emph{Findings of the Association for Computational Linguistics:
  ACL 2023}, pages 9952--9962, Toronto, Canada. Association for Computational
  Linguistics.

\bibitem[{Pandey et~al.(2025)Pandey, Budhiraja, Saha, and
  Choudhury}]{pandey-etal-2025-culturally}
Saurabh~Kumar Pandey, Harshit Budhiraja, Sougata Saha, and Monojit Choudhury.
  2025.
\newblock \href {https://aclanthology.org/2025.coling-demos.21/} {{CULTURALLY}
  {YOURS}: A reading assistant for cross-cultural content}.
\newblock In \emph{Proceedings of the 31st International Conference on
  Computational Linguistics: System Demonstrations}, pages 208--216, Abu Dhabi,
  UAE. Association for Computational Linguistics.

\bibitem[{Patra et~al.(2023)Patra, Singhal, Huang, Chi, Dong, Wei, Chaudhary,
  and Song}]{patra-etal-2023-beyond}
Barun Patra, Saksham Singhal, Shaohan Huang, Zewen Chi, Li~Dong, Furu Wei,
  Vishrav Chaudhary, and Xia Song. 2023.
\newblock \href {https://doi.org/10.18653/v1/2023.acl-long.856} {Beyond
  {E}nglish-centric bitexts for better multilingual language representation
  learning}.
\newblock In \emph{Proceedings of the 61st Annual Meeting of the Association
  for Computational Linguistics (Volume 1: Long Papers)}, pages 15354--15373,
  Toronto, Canada. Association for Computational Linguistics.

\bibitem[{Rao et~al.(2024)Rao, Yerukola, Shah, Reinecke, and
  Sap}]{rao2024normadframeworkmeasuringcultural}
Abhinav Rao, Akhila Yerukola, Vishwa Shah, Katharina Reinecke, and Maarten Sap.
  2024.
\newblock \href {https://arxiv.org/abs/2404.12464} {Normad: A framework for
  measuring the cultural adaptability of large language models}.
\newblock \emph{Preprint}, arXiv:2404.12464.

\bibitem[{Reimers and
  Gurevych(2019{\natexlab{a}})}]{reimers-gurevych-2019-sentence}
Nils Reimers and Iryna Gurevych. 2019{\natexlab{a}}.
\newblock \href {https://doi.org/10.18653/v1/D19-1410} {Sentence-{BERT}:
  Sentence embeddings using {S}iamese {BERT}-networks}.
\newblock In \emph{Proceedings of the 2019 Conference on Empirical Methods in
  Natural Language Processing and the 9th International Joint Conference on
  Natural Language Processing (EMNLP-IJCNLP)}, pages 3982--3992, Hong Kong,
  China. Association for Computational Linguistics.

\bibitem[{Reimers and
  Gurevych(2019{\natexlab{b}})}]{reimers-2019-sentence-bert}
Nils Reimers and Iryna Gurevych. 2019{\natexlab{b}}.
\newblock \href {https://arxiv.org/abs/1908.10084} {Sentence-bert: Sentence
  embeddings using siamese bert-networks}.
\newblock In \emph{Proceedings of the 2019 Conference on Empirical Methods in
  Natural Language Processing}. Association for Computational Linguistics.

\bibitem[{Shrawgi et~al.(2024)Shrawgi, Rath, Singhal, and
  Dandapat}]{shrawgi-etal-2024-uncovering}
Hari Shrawgi, Prasanjit Rath, Tushar Singhal, and Sandipan Dandapat. 2024.
\newblock \href {https://aclanthology.org/2024.eacl-long.111} {Uncovering
  stereotypes in large language models: A task complexity-based approach}.
\newblock In \emph{Proceedings of the 18th Conference of the European Chapter
  of the Association for Computational Linguistics (Volume 1: Long Papers)},
  pages 1841--1857, St. Julian{'}s, Malta. Association for Computational
  Linguistics.

\bibitem[{Song et~al.(2020)Song, Tan, Qin, Lu, and
  Liu}]{DBLP:conf/nips/Song0QLL20}
Kaitao Song, Xu~Tan, Tao Qin, Jianfeng Lu, and Tie{-}Yan Liu. 2020.
\newblock \href
  {https://proceedings.neurips.cc/paper/2020/hash/c3a690be93aa602ee2dc0ccab5b7b67e-Abstract.html}
  {Mpnet: Masked and permuted pre-training for language understanding}.
\newblock In \emph{Advances in Neural Information Processing Systems 33: Annual
  Conference on Neural Information Processing Systems 2020, NeurIPS 2020,
  December 6-12, 2020, virtual}.

\bibitem[{Storks et~al.(2019)Storks, Gao, and Chai}]{storks2019recent}
Shane Storks, Qiaozi Gao, and Joyce~Y Chai. 2019.
\newblock Recent advances in natural language inference: A survey of
  benchmarks, resources, and approaches.
\newblock \emph{arXiv preprint arXiv:1904.01172}.

\bibitem[{Villalobos et~al.(2024)Villalobos, Ho, Sevilla, Besiroglu, Heim, and
  Hobbhahn}]{villalobosposition}
Pablo Villalobos, Anson Ho, Jaime Sevilla, Tamay Besiroglu, Lennart Heim, and
  Marius Hobbhahn. 2024.
\newblock Position: Will we run out of data? limits of llm scaling based on
  human-generated data.
\newblock In \emph{Forty-first International Conference on Machine Learning
  (ICML)}.

\bibitem[{Wan et~al.(2023{\natexlab{a}})Wan, Pu, Sun, Garimella, Chang, and
  Peng}]{wan-etal-2023-kelly}
Yixin Wan, George Pu, Jiao Sun, Aparna Garimella, Kai-Wei Chang, and Nanyun
  Peng. 2023{\natexlab{a}}.
\newblock \href {https://doi.org/10.18653/v1/2023.findings-emnlp.243}
  {{``}kelly is a warm person, joseph is a role model{''}: Gender biases in
  {LLM}-generated reference letters}.
\newblock In \emph{Findings of the Association for Computational Linguistics:
  EMNLP 2023}, pages 3730--3748, Singapore. Association for Computational
  Linguistics.

\bibitem[{Wan et~al.(2023{\natexlab{b}})Wan, Zhao, Chadha, Peng, and
  Chang}]{wan-etal-2023-personalized}
Yixin Wan, Jieyu Zhao, Aman Chadha, Nanyun Peng, and Kai-Wei Chang.
  2023{\natexlab{b}}.
\newblock \href {https://doi.org/10.18653/v1/2023.findings-emnlp.648} {Are
  personalized stochastic parrots more dangerous? evaluating persona biases in
  dialogue systems}.
\newblock In \emph{Findings of the Association for Computational Linguistics:
  EMNLP 2023}, pages 9677--9705, Singapore. Association for Computational
  Linguistics.

\bibitem[{Xu et~al.(2024)Xu, Leng, Yu, and
  Xiong}]{xu2024selfpluralisingculturealignmentlarge}
Shaoyang Xu, Yongqi Leng, Linhao Yu, and Deyi Xiong. 2024.
\newblock \href {https://arxiv.org/abs/2410.12971} {Self-pluralising culture
  alignment for large language models}.
\newblock \emph{Preprint}, arXiv:2410.12971.

\bibitem[{Zhang et~al.(2025)Zhang, Zhang, Bihani, and
  Rayz}]{zhang-etal-2025-hire}
Damin Zhang, Yi~Zhang, Geetanjali Bihani, and Julia Rayz. 2025.
\newblock \href {https://aclanthology.org/2025.coling-main.529/} {Hire me or
  not? examining language model`s behavior with occupation attributes}.
\newblock In \emph{Proceedings of the 31st International Conference on
  Computational Linguistics}, pages 7891--7911, Abu Dhabi, UAE. Association for
  Computational Linguistics.

\bibitem[{Zhong et~al.(2024)Zhong, Liu, Pan, Zhang, Zhou, Liang, Wu, Lyu, Shu,
  Yu, Cao, Jiang, Chen, Li, Chen, Hu, Liu, Zhao, Xu, Dai, Zhao, Zhang, Zhao,
  Yang, Chen, Wang, Ruan, Wang, Zhao, Zhang, Ren, Qin, Chen, Li, Zidan, Jahin,
  Chen, Xia, Holmes, Zhuang, Wang, Xu, Xia, Yu, Tang, Yang, Sun, Yang, Lu,
  Wang, Chai, Li, Lu, Sun, Zhang, Ge, Hu, Zhang, Zhou, Zhang, Zhang, Liu,
  Jiang, Kong, Xiang, Ren, Liu, Jiang, Bao, Zhang, Li, Li, Liu, Shen, Sikora,
  Zhai, Zhu, and Liu}]{zhong2024evaluationopenaio1opportunities}
Tianyang Zhong, Zhengliang Liu, Yi~Pan, Yutong Zhang, Yifan Zhou, Shizhe Liang,
  Zihao Wu, Yanjun Lyu, Peng Shu, Xiaowei Yu, Chao Cao, Hanqi Jiang, Hanxu
  Chen, Yiwei Li, Junhao Chen, Huawen Hu, Yihen Liu, Huaqin Zhao, Shaochen Xu,
  Haixing Dai, Lin Zhao, Ruidong Zhang, Wei Zhao, Zhenyuan Yang, Jingyuan Chen,
  Peilong Wang, Wei Ruan, Hui Wang, Huan Zhao, Jing Zhang, Yiming Ren, Shihuan
  Qin, Tong Chen, Jiaxi Li, Arif~Hassan Zidan, Afrar Jahin, Minheng Chen,
  Sichen Xia, Jason Holmes, Yan Zhuang, Jiaqi Wang, Bochen Xu, Weiran Xia,
  Jichao Yu, Kaibo Tang, Yaxuan Yang, Bolun Sun, Tao Yang, Guoyu Lu, Xianqiao
  Wang, Lilong Chai, He~Li, Jin Lu, Lichao Sun, Xin Zhang, Bao Ge, Xintao Hu,
  Lian Zhang, Hua Zhou, Lu~Zhang, Shu Zhang, Ninghao Liu, Bei Jiang, Linglong
  Kong, Zhen Xiang, Yudan Ren, Jun Liu, Xi~Jiang, Yu~Bao, Wei Zhang, Xiang Li,
  Gang Li, Wei Liu, Dinggang Shen, Andrea Sikora, Xiaoming Zhai, Dajiang Zhu,
  and Tianming Liu. 2024.
\newblock \href {https://arxiv.org/abs/2409.18486} {Evaluation of openai o1:
  Opportunities and challenges of agi}.
\newblock \emph{Preprint}, arXiv:2409.18486.

\bibitem[{Ziems et~al.(2023)Ziems, Dwivedi-Yu, Wang, Halevy, and
  Yang}]{ziems-etal-2023-normbank}
Caleb Ziems, Jane Dwivedi-Yu, Yi-Chia Wang, Alon Halevy, and Diyi Yang. 2023.
\newblock \href {https://doi.org/10.18653/v1/2023.acl-long.429} {{N}orm{B}ank:
  A knowledge bank of situational social norms}.
\newblock In \emph{Proceedings of the 61st Annual Meeting of the Association
  for Computational Linguistics (Volume 1: Long Papers)}, pages 7756--7776,
  Toronto, Canada. Association for Computational Linguistics.

\end{thebibliography}

\clearpage
\newpage

\section*{Appendix} \label{sec:app}

\appendix


\titlecontents{section}[18pt]{\vspace{0.05em}}{\contentslabel{1.5em}}{}
{\titlerule*[0.5pc]{.}\contentspage} 


\titlecontents{table}[0pt]{\vspace{0.05em}}{\contentslabel{1em}}{}
{\titlerule*[0.5pc]{.}\contentspage} 

\startcontents[appendix] 
\section*{Table of Contents} 
\printcontents[appendix]{section}{0}{\setcounter{tocdepth}{4}} 

\startlist[appendix]{lot} 
\section*{List of Tables} 
\printlist[appendix]{lot}{}{\setcounter{tocdepth}{1}} 

\startlist[appendix]{lof} 
\section*{List of Figures} 
\printlist[appendix]{lof}{}{\setcounter{tocdepth}{1}} 

\newpage


\section{Related Work} \label{app:related-work}

\noindent \textbf{Culture-centric Research in NLP Community:} With the aim to deploy NLP technologies (e.g., LLMs) in human societies, recent research in NLP community has focused on ethics and culture centric techniques and models \cite{adilazuarda-etal-2024-towards,ziems-etal-2023-normbank,agarwal-etal-2024-ethical}. For example, \citet{DBLP:journals/corr/abs-2110-07574} have proposed Delphi, an AI system for social reasoning, \citet{hershcovich-etal-2022-challenges} characterized culture along four prominent dimensions: common ground, objectives and values, linguistic style and form, and aboutness. \citet{li2024culturellm} utilize semantic data augmentation along with WVS (World Value Survey) to train CultureLLM, \citet{DBLP:journals/corr/abs-2307-07870} look at how the values exhibited by the LLM change with changing context, \citet{10.1145/3543507.3583535} collect a corpus of cultural common-sense knowledge to help LLMs generate culturally relevant responses. \citet{alghamdi-etal-2025-aratrust} study trustworthiness of LLMs in Arabic, \citet{liu-etal-2025-whats} investigate value priority of LLMs in using realistic social scenarios, \citet{rao2024normadframeworkmeasuringcultural} develop a framework for measuring cultural adaptability of LLMs, \citet{alkhamissi-etal-2024-investigating} study cultural alignment of LLMs when prompted with low-resource language and sensitive topics vs high-resource language, \citet{cao-etal-2024-bridging} introduce cuDialog benchmark to assist dialogue agents in cultural alignment, \citet{fung-etal-2023-normsage} proposed a framework to automatically extract cultural norms from multi-lingual conversations, \citet{ammanabrolu-etal-2022-aligning} use text based games and create agents that adhere to social norms and values in an interactive setting. In this paper, it is not possible to exhaustively cover all the works, we refer the reader to a comprehensive survey on research on culture in NLP community by \citet{adilazuarda-etal-2024-towards}. Our work is inspired by the work by \citet{dwivedi-etal-2023-eticor},  where the authors create a corpus of etiquettes from major regions of the world and propose the task of Etiquette sensitivity. 


\noindent \textbf{Measuring Bias and Stereotypes:} There has been extensive research on measuring biases and stereotypes in deep models and LLMs \cite{gallegos-etal-2024-bias,shrawgi-etal-2024-uncovering}. 
In this paper, we highlight only the relevant works. \citet{10.1037/pspa0000046} propose the ABC (Agency-Belief-Communion) model of stereotypes that analyses the stereotypes associated with groups based on three dimensions, \citet{cao-etal-2022-theory} use the ABC stereotype model and a sensitivity test to discover stereotypical group-trait associations in LLMs. \citet{wan-etal-2023-personalized} identify and formulate persona bias expressed by dialogue systems while adapting to a particular persona. \citet{nadeem-etal-2021-stereoset} introduce a stereotype dataset, StereoSet, to simultaneously evaluate the language modeling ability along with stereotypical bias in LLMs. \citet{nangia-etal-2020-crows} introduce CrowS-Pairs, a stereotype dataset, to measure bias in LLMs along nine dimensions such as race, age, gender etc against historically disadvantaged groups in the U.S. \citet{wan-etal-2023-kelly} study the expression of harmful biases in LLM generated reference letter's style and content. \citet{zhang-etal-2025-hire} evaluate gender stereotypes in LLMs using occupation and hiring based question answering. \citet{do-etal-2025-aligning} make use of demographic and historical opinion data to represent the values, norms and beliefs of a persona and show the effectiveness of a new type of reasoning: Chain Of Opinions. \citet{banerjee2025navigatingculturalkaleidoscopehitchhikers} analyze cultural sensitivity in LLMs by creating a cultural harm test dataset and a culturally aligned preference dataset to restore cultural sensitivity. \citet{xu2024selfpluralisingculturealignmentlarge} create a framework called CultureSPA, for pluralistic culture alignment in LLMs. \citet{huang-yang-2023-culturally} introduce CALI (Culturally Aware Natural Language Inference) dataset to study the effects of cultural norms on language understanding task, and awareness of these norms in LLMs.  \citet{jha-etal-2023-seegull} make use of LLMs such as GPT-3 and PaLM to increase the coverage of stereotype datasets around the world. \citet{10.5555/3666122.3666316} use community engagement to collect a stereotype dataset specific to Indian context. \citet{das-etal-2023-toward} compose a Bengali dataset to evaluate gender, religious and national identity bias in NLP systems. \citet{palta-rudinger-2023-fork} present FORK, a dataset to probe models for culinary cultural biases. However, the existing metrics are not directly adaptable to our setting for measuring etiquettical bias as explained next. 


\section{\newcorpusname\ Creation Details} \label{app:corpus-creation}
This section provides a comprehensive explanation of the processes involved in the collection, preprocessing, cleaning, and filtering of the \newcorpusname\ dataset.
\subsection{Data Collection}
Etiquette data was gathered from a variety of sources, including travel websites, official cultural web pages maintained by governments, and websites featuring cultural and etiquette-related information for various countries worldwide. Additionally, we incorporated relevant content from tweets and magazine articles referencing etiquette across different cultures and countries. To enhance the dataset’s coverage, we scraped specialized web pages containing etiquette guidelines for Australian, Maori, and various African tribal cultures. A sample of data sources is provided in Appendix \ref{app:data-sources}.

\subsection{Pre Processing}
As part of the preprocessing stage, sentences with a word count of four or fewer were removed, while overly long sentences were summarized. For each region, repetitive etiquette entries were initially identified and removed using an automated python script. This was followed by a manual review to eliminate remaining instances of repeated etiquettes.  After that, all the etiquettes were carefully checked or reworded to make sure they were appropriate for the context and made sense as a whole sentence while preserving their original meaning.

\subsection{Data Labeling}
We created a list of approximately 100 characteristic words for each of the four categories of etiquette (Dining, Travel, Visits, and Business). This list will be made publicly available via a GitHub repository. Each etiquette was automatically assigned to a category if it contained one of the characteristic words. Etiquettes with none or more than one of these characteristic words were classified as ambiguous. Ambiguous etiquettes were then manually assigned to the most appropriate category based on their content. While manually assigning the etiquettes to one of the four groups, we fixed a few errors in the previous dataset. This manual assignment was done by the authors themselves and inter-annotator agreement was measured by krippendorff's $\alpha = 0.91$.\\
Each region's etiquettes were further classified into two types, indicated in the ``Label'' column as ``Positive'' or ``Negative.'' ``Positive'' etiquettes refer to behaviors that are acceptable or expected by people in a particular culture or region, whereas ``negative'' etiquettes denote behaviors that are considered unacceptable within that culture or region.

\subsection{General Etiquettes} \label{app:general-etiquette}

\noindent We categorized etiquettes that represent common facts across multiple regions, demonstrating a notable degree of similarity. By general we mean that the etiquettes are acceptable among all the regions. The labeling process involved categorizing etiquettes through common etiquette mapping, ensuring each data point was associated with its closest relation. To enhance accuracy and minimize errors, manual annotation was then conducted for each data point. This step was critical in isolating common etiquettes to ensure precise metric calculations and prevent classification errors for these data points. The distribution of data points is summarized in Table \ref{tab: General_Etiquettes_distribution}. 

\begin{figure}[h]
    \centering
    \includegraphics[scale=0.32]{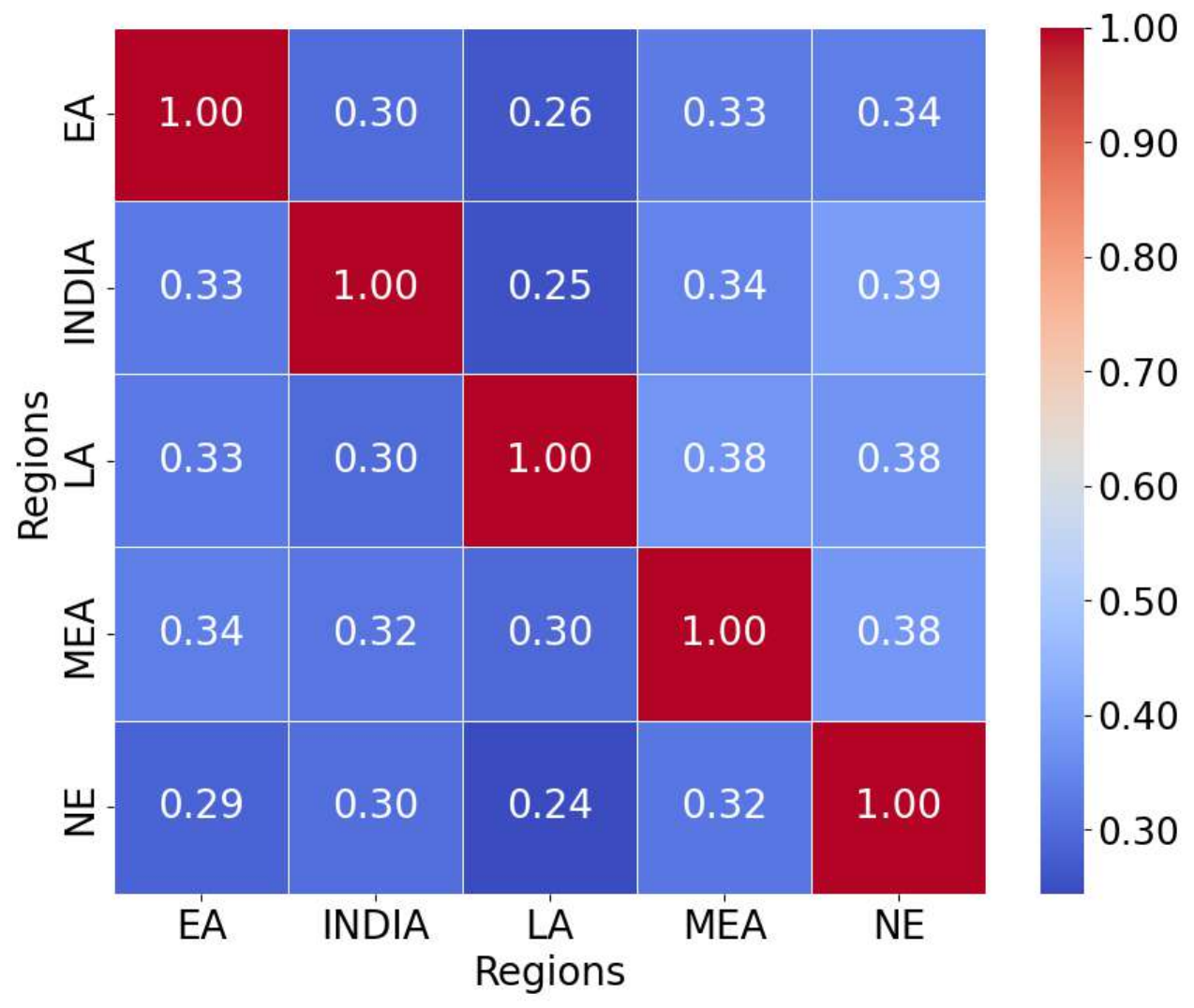}
    \vspace{-2mm}
    \caption{Region-wise Correlation for General Etiquettes}
    \label{fig:inter-region-common-corr}
    \vspace{-5mm}
\end{figure}

\noindent We also calculated the similarity matrix for only the common etiquettes of each region as described by the process in Algorithm \ref{algo:corr}.
Fig \ref{fig:inter-region-common-corr} shows that in general the distribution is nearly even throughout all regions and thus there is no particular interference of commonly accepted data types on the model results. We also avoided the use of these general etiquettes while performing tests for the metrics creation thereby removing any inconsistencies.

\begin{table}[t]
\small
\centering
\renewcommand{\arraystretch}{1}
\setlength\tabcolsep{4pt}
\begin{tabular}{lcc}
\toprule
\textbf{Region} & \textbf{\# Region Specific} & \textbf{\# General} \\
\toprule
\textbf{EA} &  8218& 1910 \\
\textbf{LA} &  5356& 1908\\
\textbf{MEA} & 9928& 3580\\
\textbf{NE} & 7488& 5162\\
\textbf{IN} &  3312& 858\\
\bottomrule
\end{tabular}
\caption{General Etiquette Distribution}
\label{tab: General_Etiquettes_distribution}
\end{table}

\section{\newcorpusname\ Region-wise Correlation} \label{app:grp-corr}

Table \ref{tab:etiquette-type-distribution-india}, Table \ref{tab:etiquette-type-distribution-ea}, Table \ref{tab:etiquette-type-distribution-la}, Table \ref{tab:etiquette-type-distribution-mea}, and Table \ref{tab:etiquette-type-distribution-ne} show the group-wise correlation between regions. 

\begin{table}[h]
\small
\centering
\renewcommand{\arraystretch}{1}
\setlength\tabcolsep{4pt}
\begin{tabular}{lcccc}
\toprule
\textbf{Region} & \textbf{\# Travel} & \textbf{\# Business} & \textbf{\# Visits} & \textbf{\# Dining} \\
\toprule
\textbf{EA} & 0.5752 & 0.5858 & 0.6221 & 0.6118 \\
\textbf{LA} & 0.8351 & 0.6667 & 0.6934 & 0.7240 \\
\textbf{MEA} & 0.8189 & 0.6784 & 0.7635 & 0.7382 \\
\textbf{NE} & 0.7067 & 0.7410 & 0.7880 & 0.6194 \\
\bottomrule
\end{tabular}
\caption{Correlation distribution for INDIA}
\label{tab:etiquette-type-distribution-india}
\end{table}

\begin{table}[h]
\small
\centering
\renewcommand{\arraystretch}{1}
\setlength\tabcolsep{4pt}
\begin{tabular}{lcccc}
\toprule
\textbf{Region} & \textbf{\# Travel} & \textbf{\# Business} & \textbf{\# Visits} & \textbf{\# Dining} \\
\toprule
\textbf{INDIA} & 0.4500 & 0.4858 & 0.5686 & 0.4137 \\
\textbf{LA} & 0.5591 & 0.5347 & 0.5336 & 0.5374 \\
\textbf{MEA} & 0.6054 & 0.5209 & 0.5662 & 0.5261 \\
\textbf{NE} & 0.5521 & 0.5500 & 0.5684 & 0.5392 \\
\bottomrule
\end{tabular}
\caption{Correlation distribution for EA}
\label{tab:etiquette-type-distribution-ea}
\end{table}

\begin{table}[h]
\small
\centering
\renewcommand{\arraystretch}{1}
\setlength\tabcolsep{3pt}
\begin{tabular}{lccccc}
\toprule
\textbf{Region} & \textbf{\# Travel} & \textbf{\# Business} & \textbf{\# Visits} & \textbf{\# Dining} \\
\toprule
\textbf{INDIA} & 0.5176 & 0.4366 & 0.4518 & 0.4490 \\
\textbf{EA} & 0.4897 & 0.4665 & 0.4766 & 0.4811 \\
\textbf{MEA} & 0.4320 & 0.4756 & 0.5501 & 0.4717 \\
\textbf{NE} & 0.5889 & 0.5689 & 0.5108 & 0.5706 \\
\bottomrule
\end{tabular}
\caption{Correlation distribution for LA}
\label{tab:etiquette-type-distribution-la}
\end{table}

\begin{table}[h]
\small
\centering
\renewcommand{\arraystretch}{1}
\setlength\tabcolsep{3pt}
\begin{tabular}{lcccc}
\toprule
\textbf{Region} & \textbf{\# Travel} & \textbf{\# Business} & \textbf{\# Visits} & \textbf{\# Dining} \\
\toprule
\textbf{INDIA} & 0.5266 & 0.4545 & 0.5309 & 0.4719 \\
\textbf{EA} & 0.5378 & 0.4645 & 0.5195 & 0.4922 \\
\textbf{LA} & 0.4621 & 0.4450 & 0.5254 & 0.4444 \\
\textbf{NE} & 0.5949 & 0.5976 & 0.6553 & 0.5721 \\
\bottomrule
\end{tabular}
\caption{Correlation distribution for MEA}
\label{tab:etiquette-type-distribution-mea}
\end{table}

\begin{table}[h]
\small
\centering
\renewcommand{\arraystretch}{1}
\setlength\tabcolsep{3pt}
\begin{tabular}{lcccc}
\toprule
\textbf{Region} & \textbf{\# Travel} & \textbf{\# Business} & \textbf{\# Visits} & \textbf{\# Dining} \\
\toprule
\textbf{INDIA} & 0.4991 & 0.4872 & 0.5899 & 0.4462 \\
\textbf{EA} & 0.5115 & 0.4819 & 0.5364 & 0.4617 \\
\textbf{LA} & 0.5235 & 0.5094 & 0.5183 & 0.4980 \\
\textbf{MEA} & 0.6112 & 0.4982 & 0.6276 & 0.4919 \\
\bottomrule
\end{tabular}
\caption{Correlation distribution for NE}
\label{tab:etiquette-type-distribution-ne}
\end{table}

\section{Prompt Templates for Various Tasks} \label{app:prompts}

\begin{tcolorbox}[width=\columnwidth,colback=gray!10,colframe=gray,sharp corners,title=Prompt used for E-Sensitivity task,coltitle=white!150!black]
    INSTRUCTION: In context of [region], answer in one word if the given cultural etiquette is (1) acceptable or (2) unacceptable. \\ \\
    USER: `[etiquette]'
\end{tcolorbox}

\begin{tcolorbox}[width=\columnwidth,colback=gray!10,colframe=gray,sharp corners,title = Prompt used for Region Identification task,coltitle=white!150!black]
    INSTRUCTION: Answer in only one word without any explanation.\\ \\
    USER: Out of the regions (1) East Asia (2) India (3) Middle East (4) South America (5) Europe, which region does the `[etiquette]' belongs to?
\end{tcolorbox}

\begin{tcolorbox}[width=\columnwidth,colback=gray!10,colframe=gray,sharp corners,title = Prompt used for Etiquette Generation task,coltitle=white!150!black]
    INSTRUCTION: You are an expert in cultural etiquettes across various countries. I will provide you with an etiquette from one region, and you need to provide, concisely in one sentence, the main corresponding etiquette from the other specified region. \\ \\
    USER: In the context of [group], Etiquette in [region1] is: `[etiquette]'.\\ \\
    ASSISTANT: The corresponding etiquette in [region2] is: 
\end{tcolorbox}


\begin{tcolorbox}[width=\columnwidth,colback=gray!10,colframe=gray,sharp corners,title=Prompt used for Incremental Option Testing,coltitle=white!150!black]
    INSTRUCTION: Select only one region from the given choices without any explanation to which the following etiquette  `[E]'  can belong to: 
    `[ List of 2 or more regions ]'\\
    USER: `[Region]'\\
\end{tcolorbox}


\section{Metric Details} \label{app:algo-metrics}

\subsection{Preference Score ($\mathbf{PS(R)})$} \label{app:sec-psr}

We calculate standard deviation to measure the difference between the model's score and the expected score as follows. It is the square root of the average of the squared difference between the model's score and the expected score. This gives us a measure of how closely the model choices reflect the actual data distribution. \\

$\sigma_{PS(R)} = \sqrt{\frac{\sum_{\mathbf{R} \in \text{Regions}} (\mathbf{PS(R)} - \mathbf{D(R)})^2}{5}}$, \\

where, $\mathbf{D(R)}$ is the percentage share of the region $\mathbf{R}'s$ data in the whole dataset

\subsection{Bias for Region Score ($\mathbf{BFS(R)}$)} \label{app:sec-bfsr}

Similar to \textbf{PS(R)}), we also calculate standard deviation to measure how extreme is the BFS distribution as follows

$\sigma_{BFS(R)} = \sqrt{\frac{\sum_{\mathbf{R} \in \text{Regions}} (\mathbf{BFS(R)} - 20)^2}{5}}$.

\subsection{Distance Metric Calculation} \label{app:sec-distancing}

Algorithm \ref{algo:distancing} describes the details of distancing metric calculation.

\begin{algorithm}[h]
\small
\renewcommand{\algorithmicrequire}{\textbf{Input:}}
\renewcommand{\algorithmicensure}{\textbf{Output:}}
\caption{Distancing Metric Calculation}\label{algo:distancing-metric}
\begin{algorithmic}
\Require 
\begin{itemize}
    \item $Q$: Total number of questions.
    \item $P$: Total number of phases (\#regions - 1).
    \item $A_{q,p} \in \{0, k, \text{abs}\}$: Action for question $q$ in phase $p$, where:
    \begin{itemize}
        \item $A_{q,p} = 0$: Correct option chosen.
        \item $A_{q,p} = k$: Incorrect option chosen (index $k$ of the option).
        \item $A_{q,p} = \text{abs}$: Abstain.
    \end{itemize}
\end{itemize}
\Ensure $D_p$: Distancing score for each phase $p$.

\State \textbf{Initialize:} $D_p \gets 0 \ \forall p \in \{0, \ldots, P-1\}$

\For{$p = 0$ \textbf{to} $P-1$}
    \State \textbf{Initialize:} $\text{Sum}_p \gets 0$
    \For{$q = 1$ \textbf{to} $Q$}
        \State \textbf{Define Score Function:}
        \[
        S(A_{q,p}) =
        \begin{cases}
        0, & \text{if } A_{q,p} = 0 \\
        -k, & \text{if } A_{q,p} = k \ (\text{incorrect option index}) \\
        1, & \text{if } A_{q,p} = \text{abs}
        \end{cases}
        \]
        \State Compute score for question $q$ in phase $p$: $s_{q,p} \gets S(A_{q,p})$
        \State Update phase sum: $\text{Sum}_p \gets \text{Sum}_p + s_{q,p}$
    \EndFor
    \State Compute average score for phase $p$: $D_p \gets \frac{\text{Sum}_p}{Q}$
\EndFor

\State \Return $\{D_0, D_1, \ldots, D_{P-1}\}$
\end{algorithmic}
\label{algo:distancing}
\end{algorithm}

\section{Discussion on Results}
\subsection{Abstentions in the E-sensitivity Task}\label{app:model-abstain}
Models abstain from answering about the acceptability of some etiquettes reasoning that the etiquette is controversial, or they are not able to understand it or in some cases they say that the etiquette is circumstantial (depending on the context) which is similar to not understanding the etiquette. We simply omit these responses from our calculations of bias. The exact count of abstentions by each model are presented in the table \ref{tab:abstentions}.
\begin{table}[htbp]
\centering
\scriptsize
\renewcommand{\arraystretch}{1} 
\setlength{\tabcolsep}{3pt} 
\begin{tabular}{@{}lccccc@{}}
\toprule
\textbf{Query} & \textbf{ChatGPT} & \textbf{Gemini} & \textbf{Llama} & \textbf{Gemma} & \textbf{Phi} \\
\midrule
Abstentions & 16 & 13 & 1986$\pm$150 & 383$\pm$35 & 144$\pm$28 \\
\bottomrule
\end{tabular}
\vspace{-3mm}
\caption{Number of abstentions for each of the five models on the E-sensitivity task.}
\label{tab:abstentions}
\vspace{-4mm}
\end{table}

\subsection{Reason for not using previous models}\label{app:reason_prev_models}
We note that previous work \cite{dwivedi-etal-2023-eticor} has used models like Falcon-40B (\url{https://huggingface.co/blog/falcon}) and Delphi \cite{DBLP:journals/corr/abs-2110-07574}. We couldn't get results on Delphi due to the inaccessibility of \href{https://delphi.allenai.org/}{Ask Delphi}. We instead decided to use more recent and efficient open-source models for our experiments.

\subsection{Tables for the Bias Score Pairwise}\label{app:bsp-results}

We present the Bias Score Pairwise for the remaining models, ChatGPT-4o, Gemini-1.5, Llama-3.1, and Gemma-2 in table \ref{tab:remaining_bspw}. 
\FloatBarrier
\begin{table*}[h]
    \centering
    \begin{minipage}{0.5\textwidth}
        \centering
        \scriptsize
         \begin{tabular}{cccccc}
            \hline
            \textbf{Region} & \multicolumn{5}{c}{\textbf{ChatGPT-4o}} \\
            ($R$/$R'$)& \textbf{NE} & \textbf{INDIA} & \textbf{EA} & \textbf{LA} & \textbf{MEA} \\
            \hline
            NE & - & \cellcolor{red!30}2.9\% & \cellcolor{green!30}48.6\% & 5.7\% & 42.9\% \\
            INDIA & 36.2\% & - & 20.3\% & \cellcolor{red!30}0.0\% & \cellcolor{green!30}43.5\% \\
            EA & 31.7\% & 4.9\% & - & \cellcolor{red!30}2.4\% & \cellcolor{green!30}61.0\% \\
            LA & \cellcolor{green!30}50.0\% & \cellcolor{red!30}2.1\% & 14.6\% & - & 33.3\% \\
            MEA & \cellcolor{green!30}57.6\% & \cellcolor{red!30}6.1\% & 24.2\% & 12.1\% & - \\
            \hline
        \end{tabular}
    \end{minipage}%
    \hfill
    \begin{minipage}{0.5\textwidth}
        \centering
        \scriptsize
        \begin{tabular}{cccccc}
            \hline
            \textbf{Region} & \multicolumn{5}{c}{\textbf{Gemini-1.5}} \\
            ($R$/$R'$)& \textbf{NE} & \textbf{INDIA} & \textbf{EA} & \textbf{LA} & \textbf{MEA} \\
            \hline
            NE & - & \cellcolor{red!30}0.0\% & \cellcolor{green!30}57.1\% & \cellcolor{red!30}0.0\% & 42.9\% \\
            INDIA & \cellcolor{green!30}67.7\% & - & 13.8\% & \cellcolor{red!30}1.5\% & 16.9\% \\
            EA & \cellcolor{green!30}68.1\% & 2.1\% & - & \cellcolor{red!30}0.0\% & 29.8\% \\
            LA & \cellcolor{green!30}64.4\% & \cellcolor{red!30}4.4\% & 6.7\% & - & 16.9\% \\
            MEA & \cellcolor{green!30}77.6\% & 2.0\% & 20.4\% & \cellcolor{red!30}0.0\% & - \\
            \hline
        \end{tabular}
    \end{minipage}

    \vspace{0.5cm}

    \begin{minipage}{0.5\textwidth}
        \centering
        \tiny
        \begin{tabular}{cccccc}
            \hline
            \textbf{Region} & \multicolumn{5}{c}{\textbf{Llama-3.1}} \\
            ($R$/$R'$)& \textbf{NE} & \textbf{INDIA} & \textbf{EA} & \textbf{LA} & \textbf{MEA} \\
            \hline
            NE & - & 33.2$\pm$2.1 & \cellcolor{green!30}45.0$\pm$2.0 & \cellcolor{red!30}5.4$\pm$1.3 & 15.7$\pm$2.2 \\
            INDIA & 37.2$\pm$1.4 & - & \cellcolor{green!30}44.8$\pm$2.2 & \cellcolor{red!30}5.1$\pm$1.0 & 12.9$\pm$1.8 \\
            EA & \cellcolor{green!30}38.5$\pm$0.9 & 31.8$\pm$1.9 & - & \cellcolor{red!30}6.6$\pm$0.4 & 23.1$\pm$1.2 \\
            LA & \cellcolor{green!30}41.4$\pm$1.7 & 14.9$\pm$0.8 & 28.0$\pm$2.5 & - & \cellcolor{red!30}15.7$\pm$1.4 \\
            MEA & 26.8$\pm$1.0 & 28.3$\pm$1.33 & \cellcolor{green!30}39.3$\pm$1.8 & \cellcolor{red!30}5.6$\pm$1.5 & - \\
            \hline
        \end{tabular}
    \end{minipage}%
    \hfill
    \begin{minipage}{0.5\textwidth}
        \centering
        \tiny
        \begin{tabular}{cccccc}
            \hline
            \textbf{Region} & \multicolumn{5}{c}{\textbf{Gemma-2}} \\
            ($R$/$R'$)& \textbf{NE} & \textbf{INDIA} & \textbf{EA} & \textbf{LA} & \textbf{MEA} \\
            \hline
            NE & - & \cellcolor{green!30}68.7$\pm$1.9 & 12.8$\pm$1.4 & \cellcolor{red!30}4.5$\pm$0.4 & 14.0$\pm$1.5 \\
            INDIA & \cellcolor{green!30}72.1$\pm$3.4 & - & 15.0$\pm$2.3 & \cellcolor{red!30}0.9$\pm$0.2 & 12.0$\pm$1.5 \\
            EA & 31.2$\pm$2.3 & \cellcolor{green!30}55.6$\pm$2.1 & - & \cellcolor{red!30}3.3$\pm$0.4 & 9.9$\pm$1.6 \\
            LA & \cellcolor{green!30}45.6$\pm$2.7 & 36.4$\pm$0.9 & \cellcolor{red!30}6.2$\pm$1.6 & - & 11.8$\pm$2.7 \\
            MEA & 31.9$\pm$2.4 & \cellcolor{green!30}59.3$\pm$1.9 & 3.8$\pm$1.6 & \cellcolor{red!30}5.0$\pm$0.5 & - \\
            \hline
        \end{tabular}
    \end{minipage}
    \caption{Result of Bias Score Pairwise for ChatGPT-4o, Gemma-2, Gemini-1.5 and Llama-3.1}
    \label{tab:remaining_bspw}
\end{table*}
\FloatBarrier

\begin{figure*}[htbp]
    \centering
    \begin{subfigure}{0.3\textwidth}
        \centering
        \includegraphics[width=\textwidth]{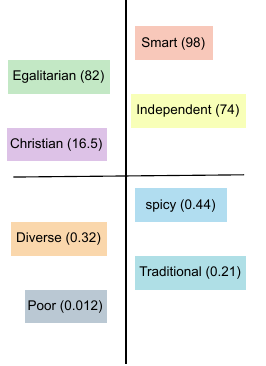}
        \caption{OR analysis of adjectives}
        \label{fig:ne_india_adj}
    \end{subfigure}
    \begin{subfigure}{0.3\textwidth}
        \centering
        \includegraphics[width=\textwidth]{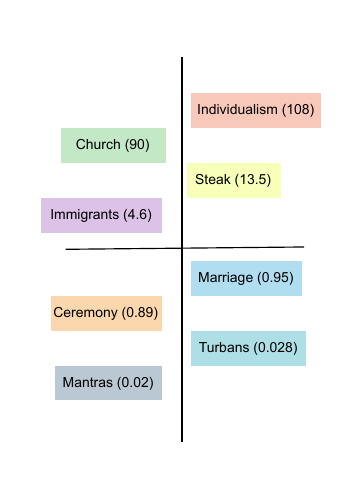}
        \caption{OR analysis of nouns}
        \label{fig:sub2}
    \end{subfigure}
    \begin{subfigure}{0.3\textwidth}
        \centering
        \includegraphics[width=\textwidth]{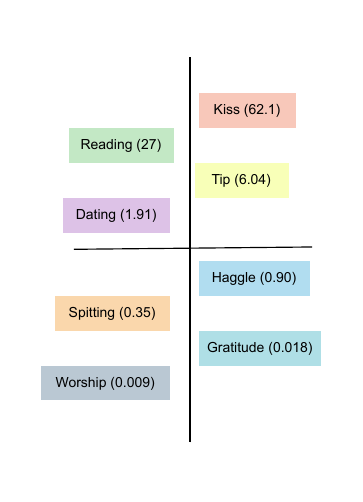}
        \caption{OR analysis of verbs}
        \label{fig:sub3}
    \end{subfigure}
    
    \caption{Odds Ratio analysis of etiquettes generated by Llama-3.1 for Europe vs India. The figure shows the words followed by their Odds Ratio.}
    \label{fig:llama31_OR}
\end{figure*}

\begin{figure*}[htbp]
    \centering
    \begin{subfigure}{0.3\textwidth}
        \centering
        \includegraphics[width=\textwidth]{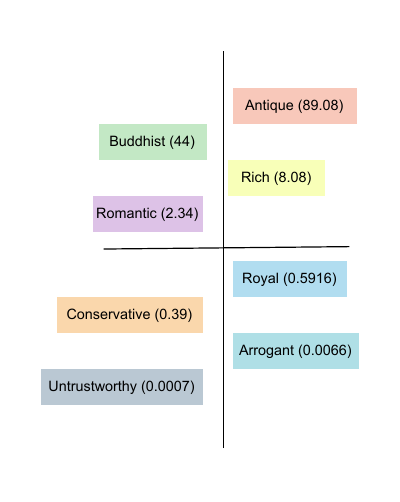}
        \caption{OR analysis of adjectives}
        \label{fig:ea_mea_adj}
    \end{subfigure}
    \begin{subfigure}{0.3\textwidth}
        \centering
        \includegraphics[width=\textwidth]{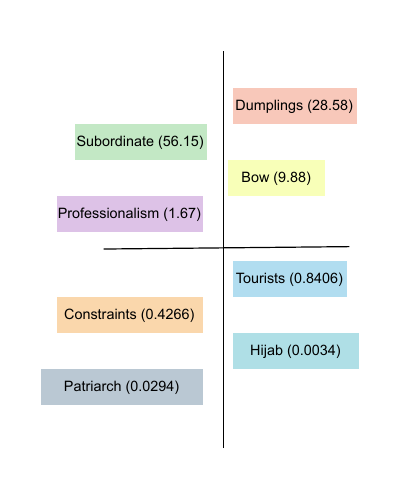}
        \caption{OR analysis of nouns}
        \label{fig:ea_mea_nouns}
    \end{subfigure}
    \begin{subfigure}{0.3\textwidth}
        \centering
        \includegraphics[width=\textwidth]{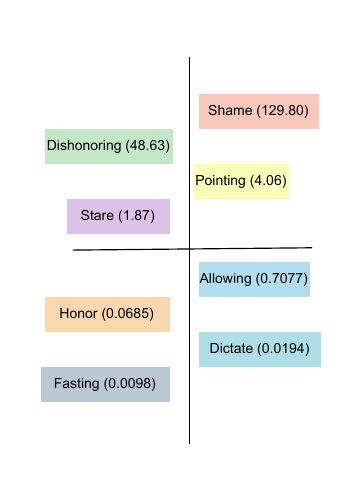}
        \caption{OR analysis of verbs}
        \label{fig:ea_mea_verbs}
    \end{subfigure}
    
    \caption{Odds Ratio analysis of etiquettes generated by Llama-3.1 for East Asia vs Middle East Africa.}
    \label{fig:llama31_OR_EA_MEA}
\end{figure*}
\begin{figure*}[htbp]
    \centering
    \begin{subfigure}{0.3\textwidth}
        \centering
        \includegraphics[width=\textwidth]{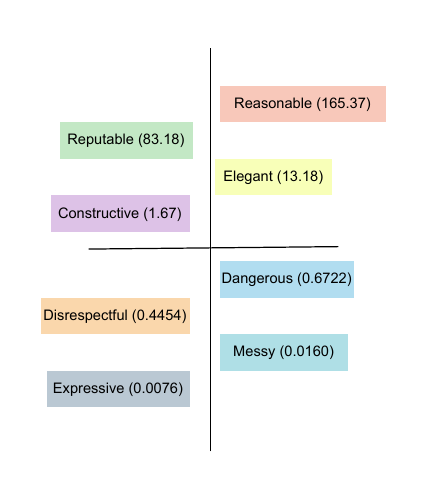}
        \caption{OR analysis of adjectives}
        \label{fig:ne_la_adj}
    \end{subfigure}
    \begin{subfigure}{0.3\textwidth}
        \centering
        \includegraphics[width=\textwidth]{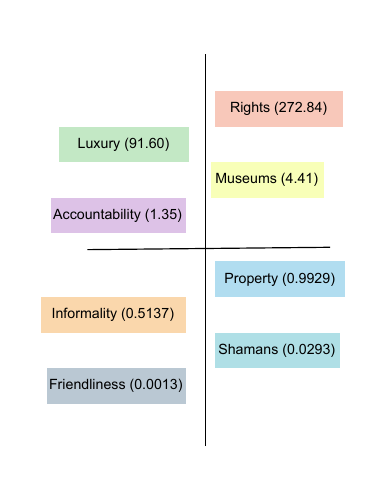}
        \caption{OR analysis of nouns}
        \label{fig:ne_la_nouns}
    \end{subfigure}
    \begin{subfigure}{0.3\textwidth}
        \centering
        \includegraphics[width=\textwidth]{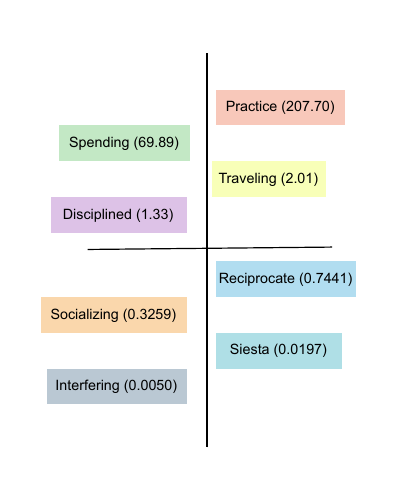}
        \caption{OR analysis of verbs}
        \label{fig:ne_la_verbs}
    \end{subfigure}
    
    \caption{Odds Ratio analysis of etiquettes generated by Phi-3.5-mini for Europe vs Latin America.}
    \label{fig:phi_or_ne_la}
\end{figure*}
\subsection{Distribution in Incremental Option Testing} \label{app:incremental-results}
\begin{figure}[H]
    \centering
    \includegraphics[scale=0.50]{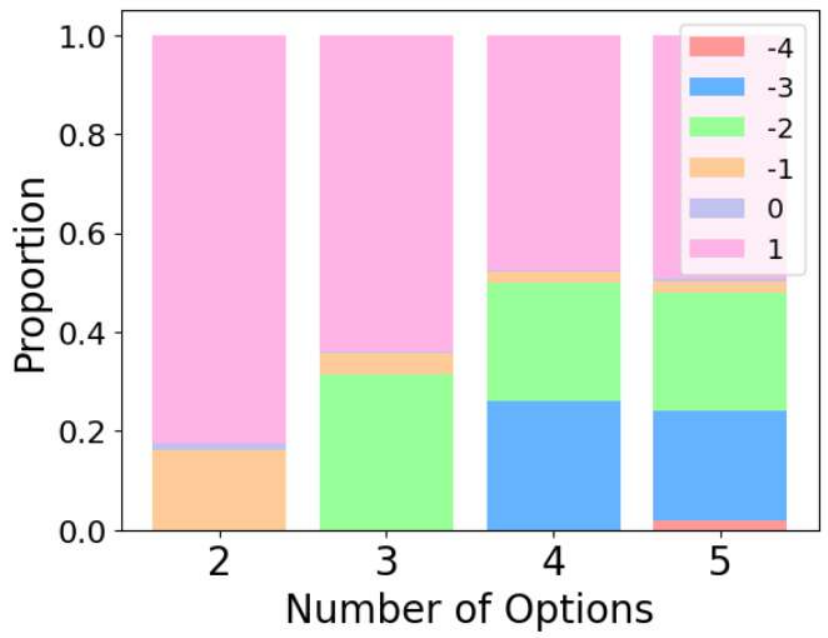}
    \caption{Distribution for Gemini Model}
    \label{fig:gemini_prop}
\end{figure}

\begin{figure}[H]
    \centering
    \includegraphics[scale=0.50]{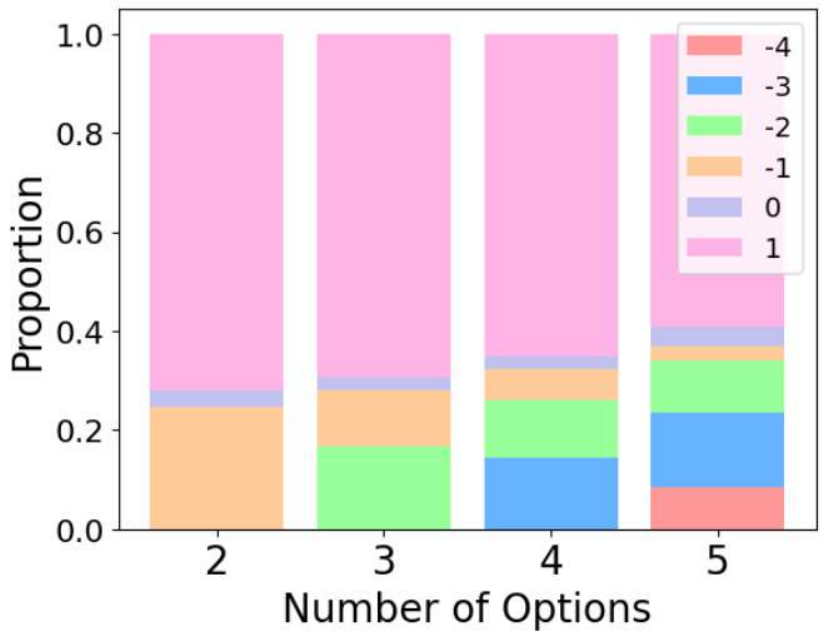}
    \caption{Distribution for Gemma Model}
    \label{fig:gemma_prop}
\end{figure}

\begin{figure}[H]
    \centering
    \includegraphics[scale=0.50]{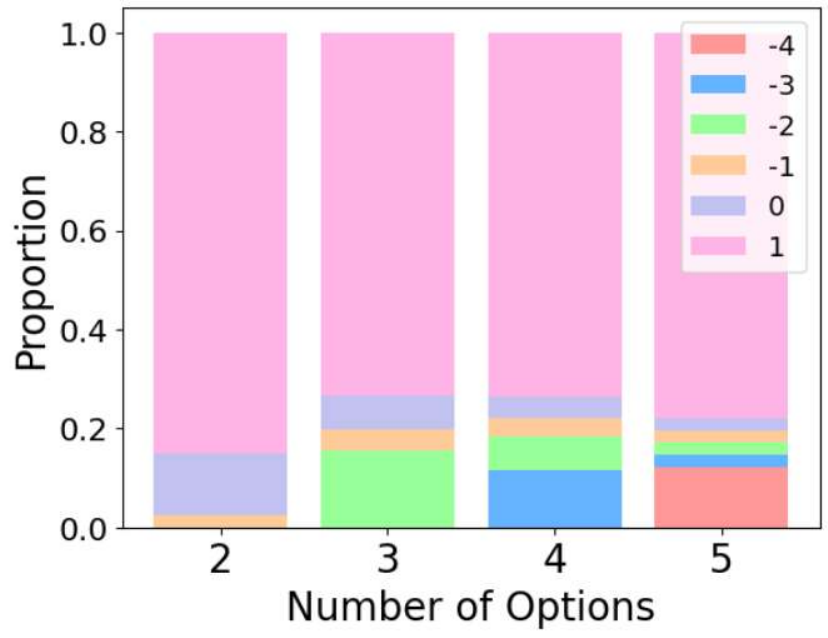}
    \caption{Distribution for ChatGPT4o Model}
    \label{fig:GPT4o_prop_prop}
\end{figure}

\begin{figure}[H]
    \centering
    \includegraphics[scale=0.50]{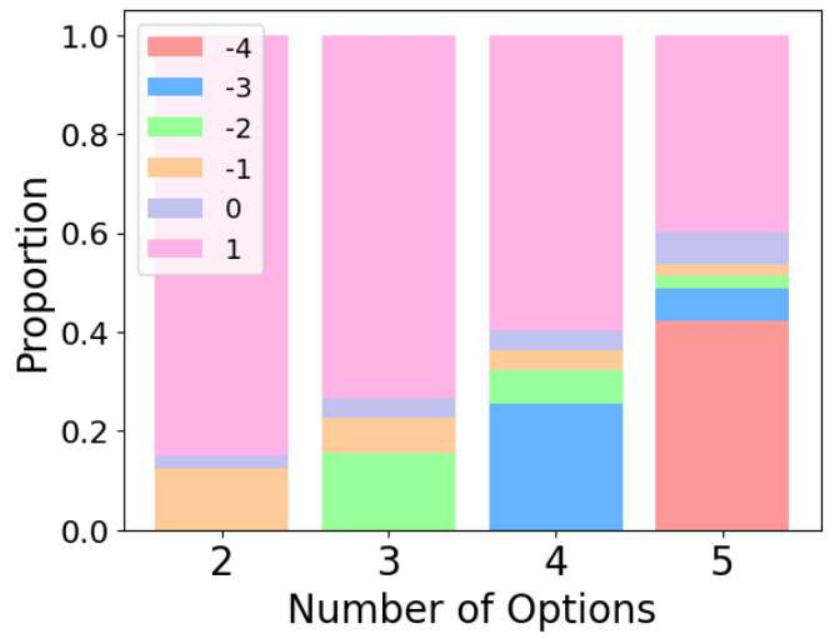}
    \caption{Distribution for Phi Model}
    \label{fig:phi_prop}
\end{figure}

\begin{figure}[H]
    \centering
    \includegraphics[scale=0.50]{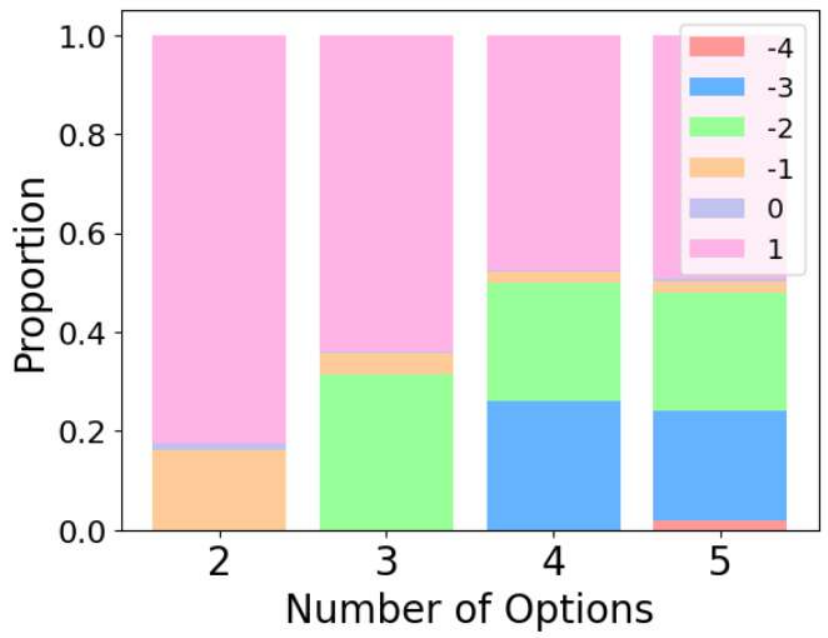}
    \caption{Distribution for Llama Model}
    \label{fig:llama_prop}
\end{figure}

Fig. \ref{fig:gemini_prop}, Fig. \ref{fig:gemma_prop}, Fig. \ref{fig:GPT4o_prop_prop}, Fig. \ref{fig:phi_prop}, Fig. \ref{fig:llama_prop} show the plots for the incremental option testing when tested via \textbf{Correct at Start Increments} method. The indicated charts are the scores for the various models throughout the process. These scores were used to calculate the metrics of distancing and accuracy over the iterations. Similar trends can be seen across models, i.e., the increase in the area of the pink portion of graphs, indicative of accuracy.

\section{Model Output Examples}
\label{sec:app-examples}

Table \ref{tab:results-examples} provides some example outputs of the models.

\begin{table*}[t]
\small
\centering
\renewcommand{\arraystretch}{1.0}
\setlength\tabcolsep{2.5pt}
\begin{tabular}{llcccccc}
\toprule
\textbf{Region} &  \begin{tabular}{c@{}} \textbf{Sentence}  \end{tabular} & \begin{tabular}{c@{}c@{}} \textbf{Gold} \\ \textbf{Label} \end{tabular} &    \textbf{ChatGPT} & \textbf{Gemini} & \textbf{Llama}  & \textbf{Gemma} & \textbf{Phi} \\
\toprule
EA & \begin{tabular}{@{}l@{}}It is customary to wash your hand before and after eating\end{tabular} & 1 & 1 & -1 & 1 & 1 &1 \\
\midrule
EA & \begin{tabular}{@{}l@{}}Feet, no matter how clean, are never placed on bed pillows. \end{tabular} & 1 & -1 & 1 & 1 & -1 & 1\\
\midrule
EA & \begin{tabular}{@{}l@{}}Public display of affection of opposite genders is fine \end{tabular} & 1 & 1 &  1& 1 &1 &1 \\
\midrule
EA & \begin{tabular}{@{}l@{}}Always touch someone's head, as it is considered \\ disrespectful.\end{tabular} & -1 & -1 & -1 & -1 & -1 & -1\\
\midrule
EA & \begin{tabular}{@{}l@{}}Blowing one's nose in public is considered good manners.\end{tabular} & -1 & -1 & -1  &  -1\\
\midrule
MEA & \begin{tabular}{@{}l@{}}If you bring a gift, expect your host to always open it in\\ front of you\end{tabular} & -1 & -1 &  -1& -1& 1& 1  \\
\midrule
MEA & \begin{tabular}{@{}l@{}}Non-Muslims are expected to disregard the fasting hours \\ in public during Ramadan.\end{tabular} & -1 & -1 &  -1& -1 &-1 &-1\\
\midrule
MEA & \begin{tabular}{@{}l@{}}Be sure your business cards are in fine shape, they are \\an extension of you as a person and must look as good \\as possible.\end{tabular} & 1 & -1 &  1& 1 & 1 & 1\\
\midrule
MEA & \begin{tabular}{@{}l@{}}Placing your right hand on your heart is a warm way to \\greet someone. 
\end{tabular} & 1 & 1 &  -1 &  1 &-1 &-1\\
\midrule
NE & \begin{tabular}{@{}l@{}}When attending a wine tasting, spit the wine into a spittoon \\if provided, especially if you are sampling multiple wines.
\end{tabular} & 1 & 1 & 1 & 1 & 1 & -1\\
\midrule
NE & \begin{tabular}{@{}l@{}}Avoid slouching or leaning back in your chair during \\the meal.\end{tabular} & 1 & 1 & 1 &  1 &1 &1\\
\midrule
NE & \begin{tabular}{@{}l@{}}Do not eat pizza with your hands.\end{tabular} & -1 & 1 & -1  & -1 & 1 & 1\\
\midrule
NE & \begin{tabular}{@{}l@{}}Participate in the conversation by interrupting others.\end{tabular} & -1 & -1 & -1 &  -1 & -1 &-1\\
\midrule
INDIA & \begin{tabular}{@{}l@{}}Never tell a girl you don't know that she is beautiful\\ or compliment on her features\end{tabular} & 1 &1 &  -1& 1 & -1 & 1\\
\midrule
INDIA & \begin{tabular}{@{}l@{}}Don't bring non-halal items into a Muslim restaurant/home.\end{tabular} & 1 & -1 &  1&  -1 & 1 & 1\\
\midrule
INDIA & \begin{tabular}{@{}l@{}}Indians are liberal when it comes to physical gesturing such \\as hand movements.\end{tabular} & -1 & 1 & -1 & 1 & -1 & 1\\
\midrule
INDIA & \begin{tabular}{@{}l@{}}India is still a very conservative nation and hugging and \\kissing are not common practices, especially with a newly \\made acquaintance\end{tabular} & 1 & 1 &  1& 1 &1 &1 \\
\midrule
INDIA & \begin{tabular}{@{}l@{}}When drinking from a water container used by others, touch \\your lips to it\end{tabular} & -1 & -1 &  -1&  -1 & -1 & 1\\
\midrule
LA & \begin{tabular}{@{}l@{}}Do not inquire about a person’s occupation or income in\\ casual conversation, although that may be inquired of you.\end{tabular} & 1 & 1 &  -1&  1 & -1 & 1\\
\midrule
LA & \begin{tabular}{@{}l@{}}In the workplace, colleagues of similar status may call each\\ other by their first names.\end{tabular} & 1 & 1 & 1 & 1 & 1& 1 \\
\midrule
LA & \begin{tabular}{@{}l@{}}Always speak with your hands in your pockets, it is \\considered polite\end{tabular} & -1 & -1 &  -1&  -1 & -1 &-1\\ \midrule
\end{tabular}
\caption{Some Examples of Etiquette's and their corresponding zero shot results on the E-sensitivity task.}
\label{tab:results-examples}
\end{table*}

\section{Sample Data Sources}\label{app:data-sources}
We present some sample data sources from where we scrapped our data. A complete list of data sources along with the data will be provided in the GitHub repository after acceptance.
\begin{itemize}
    \item \url{https://guide.culturecrossing.net/basics_business_student_details.php?Id=29&CID=143}
\item \url{https://www.britannica.com/place/Russia/Daily-life-and-social-customs}
\item \url{https://guide.culturecrossing.net/basics_business_student_details.php?Id=10&CID=148}
\item \url{https://www.entriva.com/en/blog/tanzania-tradition-culture/#:~:text=Always%20eat%20with%20your%20right,it%20is%20welcome%20and%20appreciated.}
\item \url{https://guide.culturecrossing.net/basics_business_student_details.php?Id=29&CID=43}
\item \url{https://guide.culturecrossing.net/basics_business_student_details.php?Id=23&CID=80}
\item \url{https://guide.culturecrossing.net/basics_business_student_details.php?Id=27&CID=171}
\item \url{https://guide.culturecrossing.net/basics_business_student_details.php?Id=15&CID=107}
\item \url{https://culturalatlas.sbs.com.au/australian-culture/australian-culture-business-culture}
\item \url{https://guide.culturecrossing.net/basics_business_student_details.php?Id=11&CID=13}
\item \url{http://web.sut.ac.th/cia/2017/CulturalEtiquette/ChinaCulturalEtiquette.pdf}
\item \url{https://culturalatlas.sbs.com.au/russian-culture/russian-culture-communication}
\item \url{https://culturalatlas.sbs.com.au/chinese-culture/chinese-culture-business-culture}
\item \url{https://guide.culturecrossing.net/basics_business_student_details.php?Id=20&CID=43}
\item \url{https://asianabsolute.co.uk/blog/russian-business-etiquette-dos-and-donts/}
\item \url{http://www.namibia-travel-guide.com/bradt_guide.asp?bradt=1052}
\item \url{https://www.24hourtranslation.com/algeria-culture-customs-and-social-etiquette.html}
\item \url{https://guide.culturecrossing.net/basics_business_student_details.php?Id=23&CID=107}
\item \url{https://guide.culturecrossing.net/basics_business_student_details.php?Id=7&CID=123}
\item \url{https://guide.culturecrossing.net/basics_business_student_details.php?Id=20&CID=123}
\end{itemize}


\end{document}